\documentclass[a4paper,review]{cas-sc}
\usepackage[left]{lineno}

\usepackage[authoryear]{natbib}
\usepackage{multirow}

\usepackage{tabularx}
\usepackage{caption}

\usepackage{soul}  
\usepackage{xcolor}  

\newif\ifshowrevone  
\showrevonefalse  
\newcommand{\change}[1]{\ifshowrevone\hl{#1}\else#1\fi}  
\newcommand{\added}[1]{\ifshowrevone\textcolor{blue}{#1}\else#1\fi}

\newif\ifshowcitehighlight
\showcitehighlightfalse 
\newcommand{\newcitep}[1]{\ifshowcitehighlight\colorbox{blue!20}{\citep{#1}}\else\citep{#1}\fi}

\begin{document}

\shorttitle{}

\begin{center}
\LARGE\textbf{Enhancing Geo-localization for Crowdsourced Flood Imagery via LLM-Guided Attention}
\end{center}

\begin{center}
\large
Fengyi Xu$^{a}$,
Jun Ma$^{a,b,*}$,
Waishan Qiu$^{a}$,
Cui Guo$^{a,b}$,
Jack C.P. Cheng$^{c}$
\end{center}

\begin{center}
\small
$^{a}$ Department of Urban Planning and Design, The University of Hong Kong, Hong Kong SAR, China\\[2pt]
$^{b}$ Urban Systems Institute, The University of Hong Kong, Hong Kong SAR, China\\[2pt]
$^{c}$ Department of Civil and Environmental Engineering, The Hong Kong University of Science and Technology, \\
Hong Kong SAR, China
\end{center}

\begin{center}
\small
$^{*}$\textit{Corresponding author}: \href{mailto:junma@hku.hk}{junma@hku.hk}\\[2pt]
Contact: \href{mailto:fengyixu_arch@connect.hku.hk}{fengyixu\_arch@connect.hku.hk} (Fengyi Xu)
\end{center}

\vspace{1em}

\noindent
\textbf{Abstract}\\
Crowdsourced street-view imagery from social media provides valuable real-time visual evidence of urban flooding and other crisis events, yet it often lacks reliable geographic metadata for emergency response. Existing image geo-localization approaches, also known as Visual Place Recognition (VPR) models, exhibit substantial performance degradation when applied to such imagery due to visual distortions and domain shifts inherent in cross-source scenarios. This paper presents VPR-AttLLM, a model-agnostic framework that integrates the semantic reasoning and geospatial knowledge of Large Language Models (LLMs) into established VPR pipelines through attention-guided descriptor enhancement. By leveraging LLMs to identify location-informative regions within the city context and suppress transient visual noise, VPR-AttLLM improves retrieval performance without requiring model retraining or additional data. \change{To evaluate the framework, we conduct comprehensive testing across two morphologically distinct urban environments: San Francisco and Hong Kong. The evaluation utilizes established query sets, synthetic flooding scenarios, and real social media flood images integrated into the San Francisco benchmark, alongside a newly curated Hong Kong dataset.} Integrating VPR-AttLLM with three state-of-the-art VPR models—CosPlace, EigenPlaces, and SALAD—consistently improves recall performance, yielding relative gains typically between 1–3\% and reaching up to 8\% on the most challenging real flood imagery. Beyond measurable gains in retrieval accuracy, this study \change{demonstrates a robust} pipeline for LLM-guided multimodal fusion in visual retrieval systems. By embedding principles from urban perception theory into attention mechanisms, VPR-AttLLM bridges human-like spatial reasoning with modern VPR architectures. Its plug-and-play design, strong cross-source robustness, and interpretability highlight its potential for scalable urban monitoring and rapid geo-localization of crowdsourced crisis imagery \change{in heavily urbanized environments}.


\section{Introduction}
Street View Imagery (SVI) has become an essential source for urban observation and analysis because it enables detailed, ground-level perspectives on built environments \citep{gebruUsingDeepLearning2017, naikComputerVisionUncovers2017}. While platforms such as Google Street View (GSV) provide standardized and extensive spatial coverage, they lack high-frequency temporal updates and cannot capture ephemeral or crisis-related changes in urban scenes—such as flash floods and hurricanes. Consequently, crowdsourced street‑view imagery (SVI) collected through social media and citizen‑reporting platforms has emerged as a valuable complement for timely urban sensing and monitoring \citep{goodchildCitizensSensorsWorld2007, biljeckiStreetViewImagery2021}, enhancing cities’ capacity to assess and respond to rapidly evolving situations, particularly during emergencies \citep{qinEnhancingUrbanResilience2025}.

\begin{figure}[pos=htbp]  
    \centering
    \includegraphics[width=1\textwidth]{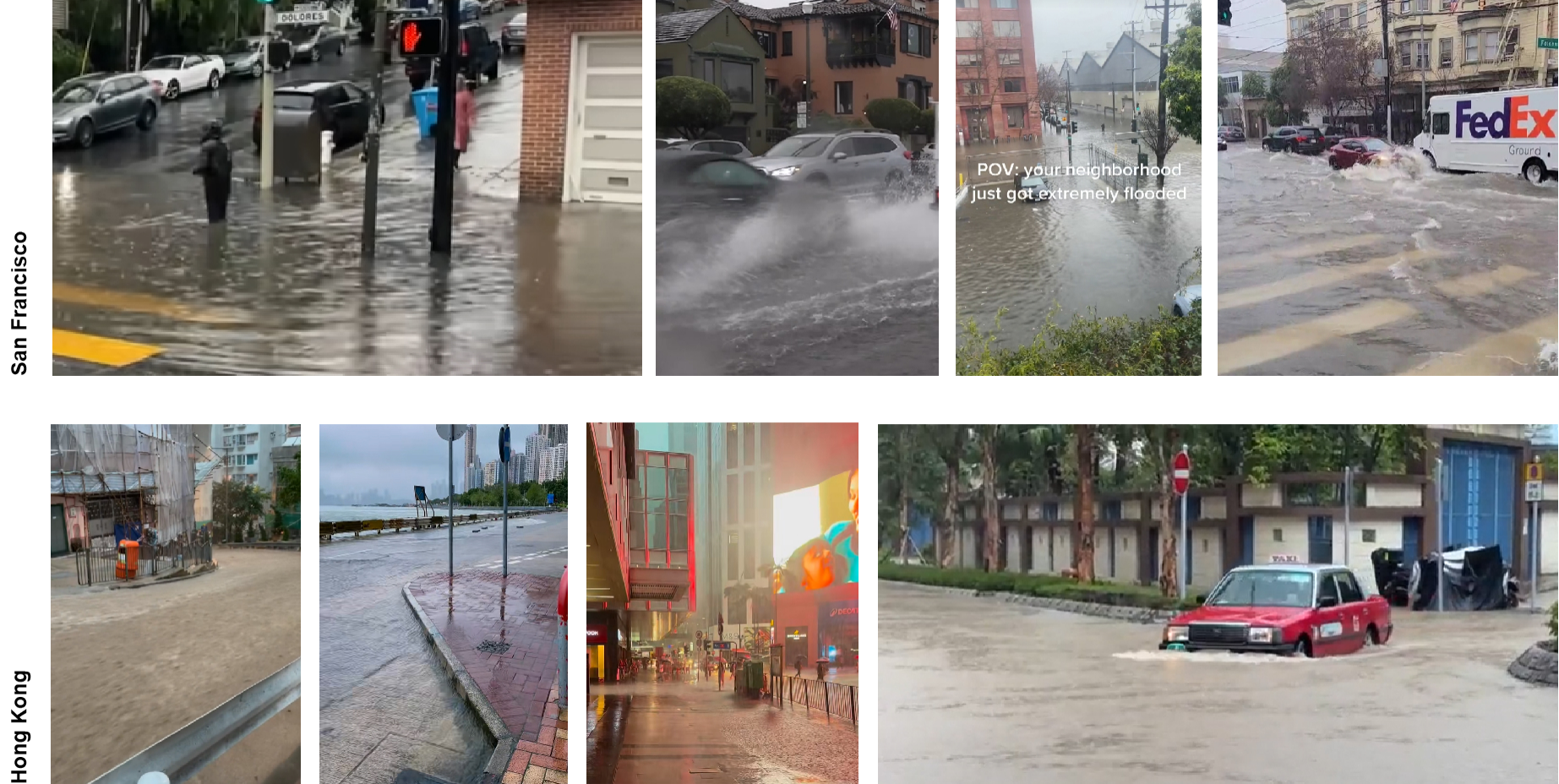} 
    \caption{Examples of flood-related street view imagery shared on social media. Top: San Francisco; bottom: Hong Kong. Transient water patterns, reflections, and occlusions substantially alter the street appearance, posing severe challenges for visual place recognition systems trained under normal-weather distributions.} 
    \label{fig:intro_svi}  
\end{figure}

Accurately locating crowdsourced street images is vital for integrating them into spatial decision systems for emergency response and urban resilience planning. This remains challenging because most contributors omit precise geotags for privacy or lack of geospatial literacy for precise locations \citep{croitoruGeosocialGaugeSystem2013, tavraRoleCrowdsourcingSocial2021}. Although many cities have deployed official reporting platforms such as New York’s 311 system, persistent reporting friction and limited public awareness constrain participation, leading to biased and incomplete representations of urban issues \citep{kontokostaBiasSmartCity2021, boxerEstimatingReportingBias2025}. During disruptive events such as sudden flooding, citizens often share unstructured photos or videos on social media—data that lack geographic metadata yet offer rich, real‑time visual evidence of urban disruptions (Figure~\ref{fig:intro_svi}). Such imagery complements formal reporting systems by capturing transient, on‑the‑ground scenes that enhance situation awareness. However, current municipal workflows still depend on manual verification, or cross‑referencing by local agencies and journalists, which delays response and limits scalability. Furthermore, this lack of real-time ground verification disconnects dynamic Urban Digital Twins from the physical reality they aim to simulate, rendering them less effective during rapidly unfolding crises.

Geo-localization, also known as Visual Place Recognition (VPR), aims to infer the geographic location of an image by matching it to a geo-tagged reference database. VPR lies at the intersection of computer vision, urban morphology, and environmental perception. Foundational urban perception theories, such as the “Image of the City” \citep{lynchImageCity2008}, "Scale and Psychologies of Space" \citep{montelloScaleMultiplePsychologies1993}, highlight the role of legibility and structural coherence in spatial recognition, while later computational works have quantified discriminative urban visual elements—demonstrating, for instance, “What makes Paris look like Paris?” \citep{doerschWhatMakesParis2012}. 

Modern VPR approaches fall broadly into two categories. Retrieval-based methods treat VPR as an image retrieval problem, building city-specific reference databases and retrieving the most similar reference image for a given query—exemplified by NetVLAD \citep{arandjelovicNetVLADCNNArchitecture2016}, CosPlace \citep{bertonRethinkingVisualGeolocalization2022a}, and more recent SALAD \citep{izquierdoOptimalTransportAggregation2024}. Classification-based VPR models instead predict a coarse geographic region or city category, allowing global coverage at lower spatial resolution, as demonstrated in PlaNet \citep{weyandPlaNetPhotoGeolocation2016}, GeoCLIP \citep{cepedaGeoCLIPClipinspiredAlignment2023}, and PIGEON \citep{haasPIGEONPredictingImage2024}. With the progression of deep neural networks, both approaches have achieved significant progress by employing CNN-based backbones (e.g., VGG16, ResNet), vision transformers (e.g., DINO), and multimodal representations (e.g., CLIP). 

Despite these advances, robust geo-localization in real-world, crowdsourced SVI remains highly challenging. Even models achieving strong performance on established benchmarks (e.g., SF-XL \citep{bertonRethinkingVisualGeolocalization2022a}, Pittsburgh-30k \citep{toriiVisualPlaceRecognition2015a}, Tokyo24/7 \newcitep{torii247Place2018}) exhibit significant accuracy drops when confronted with distribution shifts, such as extreme weather-induced visual distortions (e.g., flooding, heavy rain) or cross-source variations in data capture devices and environmental conditions. These challenges exemplify a fundamental problem in discriminative AI models: \textbf{undergeneralization}, where model performance degrades considerably when inputs deviate from training distributions \citep{ilievskiAligningGeneralizationHumans2025}. This issue proves particularly acute in urban geo-localization for two reasons: (i) extreme weather events generate inherently limited training data due to their rarity, and (ii) the diversity of potential distribution shifts makes scenario-specific retraining computationally prohibitive. Addressing undergeneralization in VPR therefore requires training-free approaches that can dynamically leverage complementary reasoning mechanisms to enhance robustness under challenging conditions without requiring extensive retraining or domain-specific data collection.

The emergence of multimodal Large Language Models (LLMs)—such as GPT-4o, Gemini 2.5, and Qwen-VL—creates new opportunities for resolving these limitations. Beyond text understanding, modern LLMs demonstrate sophisticated scene reasoning and spatial awareness derived from large-scale cross-modal training. They can identify and interpret urban components (e.g., building façades, vegetation, signage, or street furniture) and contextual cues that human observers use to infer place identity. Recent project like GeoIntel \citep{atiilla2024geointel} show that LLMs can perform coarse geo-reasoning and recognize landmark-rich scenes; however, without integration into established retrieval-based pipelines, such models often lack street-level precision for city-scale VPR tasks.

To address this gap, this paper proposes a \textbf{VPR enhancement framework with LLM-guided attention (VPR-AttLLM)}. By treating crowdsourced imagery as a sensor network, this framework serves as a robust data assimilation mechanism for Urban Digital Twins, leveraging LLMs’ semantic reasoning and general geo-knowledge to enhance retrieval-based VPR pipelines under challenging conditions. The central idea is to guide existing pre-trained VPR models with LLM-derived attention map extracted from query images, thereby emphasizing salient regions containing strong location cues (e.g., architectural structure, signage, or horizon geometry) and suppressing noisy regions like flooding street surface (Figure~\ref{fig:intro_svi}). This integration does not require additional training or supporting data, making it model-agnostic and computationally efficient.

Comprehensive experiments demonstrate that the proposed VPR-AttLLM consistently enhances multiple pre-trained VPR models across diverse VPR tests. When applied to CosPlace and EigenPlaces, VPR-AttLLM yields improvements of up to 8\% on real flood imagery in the newly constructed HK\_Flood dataset and around 4\% on the SF\_Flood set from San Francisco, while maintaining 1–3\% gains under other synthetic flooding and domain-shifted Mapillary scenarios. Notably, the framework demonstrates positive transfer even to the transformer-based SALAD model with a DINOv2 backbone which already achieves strong baseline performance. These results collectively highlight the framework’s ability to reinforce retrieval robustness under severe visual perturbations and cross-source appearance shifts, confirming the generality of LLM-guided attention integration framework.

In summary, this study contributes a novel and generalizable approach for improving VPR robustness under severe appearance or domain shifts by harnessing the semantic reasoning capability and geo-knowledge of large language models. Beyond quantitative gains, the proposed framework highlights the potential for fusing LLM-driven perception with visual retrieval pipelines for resilient urban sensing. The remainder of this paper is organized as follows. Section~\ref{sec:relatedwork} reviews related work on VPR models and LLM-assisted geo-localization. Section~\ref{sec:method} details the proposed VPR-AttLLM framework. Section~\ref{sec:experiments} describes the implementation details, the model performance, and comparative findings. Section~\ref{sec:discussion} discusses implications for multimodal urban sensing, while Section~\ref{sec:conclusion} concludes the study.

\section{Related Work}  
\label{sec:relatedwork}

\subsection{Visual Place Recognition (VPR) Models}
\label{subsec:vprmodels}

Visual Place Recognition (VPR) is commonly formulated as a large-scale image retrieval task, aiming to match a query image to its corresponding geo-referenced counterpart within a database \citep{schindlerCityscaleLocationRecognition2007, torii247Place2018}. The primary challenge is to produce compact, discriminative global descriptors that remain robust to changes in viewpoint, illumination, and environmental conditions \citep{zaffarVPRBenchOpenSourceVisual2021, liPlaceRecognitionMeet2025}.

Earlier VPR approaches leveraged handcrafted local descriptors (e.g., SIFT, VLAD), but were later superseded by deep convolutional architectures capable of learning aggregation functions end-to-end. A seminal work, NetVLAD, introduced a differentiable VLAD layer on top of CNN backbones to jointly optimize feature representation and place discrimination \citep{arandjelovicNetVLADCNNArchitecture2016}. Among subsequent CNN-based methods, a major line of progress centered on learnable pooling functions. The Generalized Mean (GeM) pooling extended conventional max- and average-pooling by learning per-channel exponents, thereby enabling more discriminative global descriptors \citep{radenovicFinetuningCNNImage2018}. Building upon this formulation, classification-based VPR frameworks such as CosPlace and EigenPlaces design geo-balanced training strategies that explicitly exploit the strengths of GeM pooling to improve viewpoint and illumination invariance \citep{bertonRethinkingVisualGeolocalization2022a, bertonEigenPlacesTrainingViewpoint2023}. These works demonstrate how explicit sampling and hierarchical classification amplify the representation capacity of GeM-aggregated descriptors for robust CNN-based VPR.  

A parallel direction emphasizes improving aggregation itself through more expressive clustering or token-based formulations. MixVPR introduced feature mixing for better local–global integration \citep{ali-beyMixVPRFeatureMixing2023}, while optimal-transport–based architectures such as SALAD extend this concept by learning score-aware cluster aggregation \citep{izquierdoOptimalTransportAggregation2024}. Transformer-based models, including DINO-ViT and TransVPR, further increasingly adopted with self-attention mechanisms to capture long-range spatial dependencies and offer token-level feature reasoning \citep{hauslerPatchNetVLADMultiscaleFusion2021, wangTransVPRTransformerbasedPlace2022}. Despite these advances, most current methods remain domain-specialized and exhibit performance degradation under severe environmental or cross-source shifts \citep{gargRevisitAnythingVisual2024}. 

Beyond architectural innovations in feature extraction and aggregation, complementary post-processing strategies offer performance improvements without requiring model retraining. Among these approaches, \textit{Query Expansion} (QE) \citep{chumTotalRecallAutomatic2007} has emerged as a commonly adopted technique in VPR pipelines. Average Query Expansion (AQE) refines the query descriptor by incorporating weighted averages of top-$k$ retrieved database features, with a tunable parameter $\alpha$ controlling the balance between the original query and retrieved candidates. This simple strategy has demonstrated consistent recall improvements with minimal computational overhead across various VPR benchmarks \citep{radenovicRevisitingOxfordParis2018, gordoAttentionbasedQueryExpansion2020}. However, existing post-processing methods have not been systematically evaluated under challenging cross-domain or extreme weather scenarios with severe appearance variations, and semantic-guided query augmentation strategies that leverage high-level scene understanding remain largely unexplored.
Addressing these limitations motivates our proposed \textbf{VPR-AttLLM} framework, which introduces language model guided attention modules for both GeM pooling and cluster-based aggregation, thereby unifying these two architectural paradigms under a single interpretive mechanism. By incorporating semantic reasoning into the feature aggregation process, our approach extends beyond conventional post-processing techniques to enable more adaptive query augmentation across diverse environmental conditions.

\subsection{Integrating LLMs into Street-View Perception and Geo-Localization} 
\label{subsec:llmgeoloc}

\change{To address the need for high-level semantic reasoning in visually challenging environments, recent efforts have increasingly explored LLMs. This semantic capability is especially critical in the context of extreme weather and disaster management, where accurate, fine-grained geo-localization of multi-source information—such as crowdsourced flood imagery—is essential for emergency response. Despite this pressing need, the integration of LLMs into spatial reasoning and geo-localization tasks has predominantly remained at a coarse resolution. In recent years, researchers have widely utilized GeoGuessr-like experiments to benchmark the geographic knowledge of LLMs, often prompting them to infer countries or cities from street-view imagery} \citep{wangLLMGeoBenchmarkingLarge2024, atiilla2024geointel}. \change{Even when applied to disaster scenarios, localization performance remains broad; for example,} \citet{yinLLMenhancedDisasterGeolocalization2025} \change{achieved over 60\% accuracy within a 10~km radius for Hurricane Harvey imagery, a resolution that is highly valuable, yet still too coarse for targeted, street-level urban disaster response.}

\change{This limitation stems from a fundamental computational bottleneck in applying LLMs to visual retrieval. To circumvent the prohibitive inference costs associated with querying vast visual databases, current frameworks predominantly adopt a retrieval-free paradigm in which models are prompted to infer locations directly from a single image. While avoiding these large-scale lookups reduces computational overhead, it inherently restricts localization to coarse city or country levels, precluding LLMs from serving as precise, street-level matching agents. Consequently, in the pursuit of fine-grained disaster geolocation, current LLM adoption centers mainly on text preprocessing, leaving their direct integration into visual place recognition (VPR) at a coarse level} \citep{xuLargeLanguageModel2025}. \change{As a result, existing LLM-involved frameworks remain impractical for fine-grained urban sensing tasks where highly accurate VPR is essential.}

In parallel, LLMs have demonstrated strong capabilities in interpreting localized urban scenes and performing multimodal classification tasks. They can infer building ages from street-view imagery \citep{zengZeroshotBuildingAge2024a}, estimate inundation depth from flooding images \citep{lyuAssessingLargeMultimodal2025}, or identify improper on-street waste placement \citep{zhangMonitoringStreetlevelImproper2025} through zero- or few-shot reasoning. However, these applications primarily exploit LLMs as semantic decoders that extract or classify local visual attributes within individual images, rather than as spatially contextual components that influence global feature organization. Additionally, although several relevant experiments have been conducted in LLM's spatial cognition \citep{fengCityGPTEmpoweringUrban2025} or geo-knowledge \citep{manviGeoLLMExtractingGeospatial2024a}, existing tests are still constrained in local scenario inference like demographic indicator prediction. Consequently, while LLMs excel at local semantic interpretation, their potential to guide spatial reasoning or retrieval-level representation in VPR pipelines remains unexplored or underutilized.

Bridging these two gaps—retrieval inefficiency and local-only perception—motivates our proposed \textbf{VPR-AttLLM} framework. Instead of treating LLMs as independent classifiers or scene interpreters, VPR-AttLLM integrates their geographical knowledge and semantic reasoning directly into the global feature aggregation stage of VPR models. By introducing LLM-guided attention modules for both pooling- and cluster-based aggregation, the framework allows LLMs to modulate the weighting of visually informative regions across the entire retrieval procedure. This integration conceptually transforms LLMs from localized predictors into global reasoning agents that enhance cross-condition robustness and fine-resolution geo-localization within existing vision architectures.

\section{Methodology}  
\label{sec:method}

LLMs are capable of inferring the geographic context of a given street view image (SVI) by reasoning about its visual cues and recognizing scene elements that may reveal the underlying location. However, not all image regions contribute equally to this inference. For example, buildings generally carry more discriminative information for localization than roads or skies, and a visually distinctive landmark may provide stronger spatial cues than ordinary residential structures. Conventional VPR models, even when trained on large and diverse datasets, often lack the contextual awareness to assign such importance adaptively—particularly when encountering unseen cities with distinct architectural styles. Moreover, under challenging environmental conditions such as flooding or heavy rain, when query images differ drastically in appearance from the reference database, common VPR models tend to suffer performance degradation due to diminished visual clarity or occlusion. In these cases, an LLM’s capability to reason about spatial salience and refocus on informative regions offers a potential pathway to enhance VPR robustness.

Motivated by this insight, we propose the LLM‑Att framework, an integrated system that leverages an LLM’s broad geo‑semantic knowledge to guide attention in existing pretrained VPR models. The framework consists of two key components:
(1) an LLM attention generation module, which guides the LLM to produce spatial attention maps corresponding to salient image regions; and
(2) an LLM attention integration module, which fuses external attention maps into various pretrained VPR architectures for inference.
Together, these components enable the VPR models to adaptively weight image features according to semantic relevance, improving retrieval performance in challenging queries.

\subsection{LLM Attention Generation for SVI Query}
\label{sec:att_gen}

In VPR, assigning greater emphasis to informative regions and down-weighting less relevant or noisy areas is crucial for accurately describing a SVI and matching it with images in a reference database. This prioritization aligns with both empirical findings from VPR model training and established principles from urban perception theory. Training dynamics of state-of-the-art VPR models have consistently demonstrated that permanent infrastructure, such as building facades and established vegetation, contributes more reliably to place recognition than transient or generic elements like sky, road surfaces, or moving vehicles. This observation resonates with human spatial cognition patterns documented in environmental psychology literature. However, under challenging visual conditions—such as flooding, extreme illumination shifts, or severe occlusions commonly encountered in crowdsourced social media imagery—the learned feature representations may inadequately capture the optimal region contribution patterns, as models trained on standard street-view datasets have not encountered such distributional shifts during training.

\begin{figure}[pos=htbp]  
    \centering
    \includegraphics[width=\textwidth]{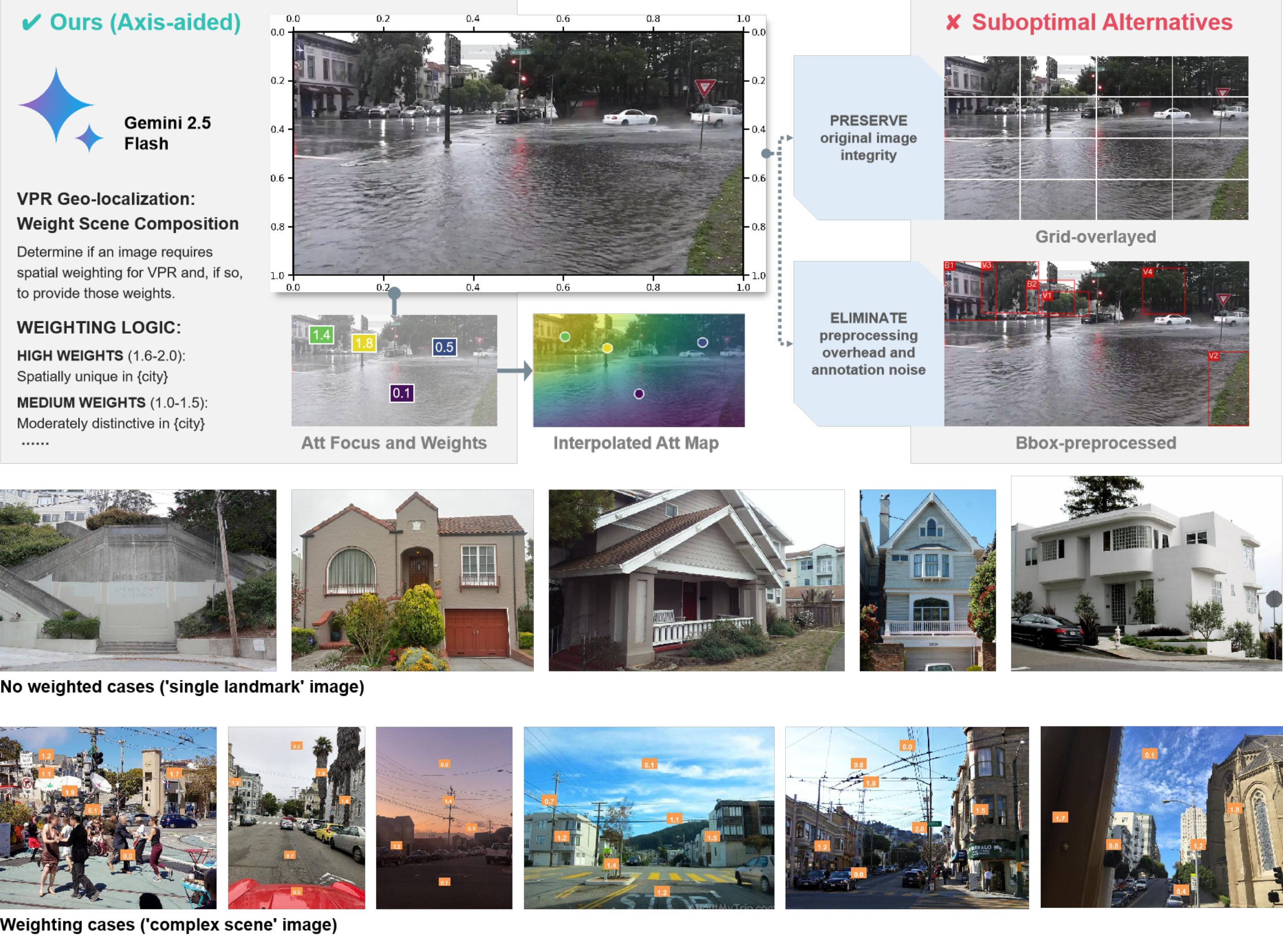} 
    \caption{LLM-guided spatial attention weighting for Visual Place Recognition. Axis-based coordinate prompting (top left) guides Gemini 2.5 Flash to generate spatial attention maps with weight values for SF-XL queries. Alternative visual prompting formats are shown for comparison (top right). High weights (1.6--2.0) are assigned to spatially unique landmarks; medium weights (1.0--1.5) to moderately distinctive features. Middle row: single-landmark scenes requiring no weighting. Bottom row: complex urban scenes with weighted attention (shown in orange tags) on distinctive architectural elements (clock towers, ornamental facades) versus generic features (sky, pavement).}
    \label{fig:method_attgen}  
\end{figure}

Building upon this foundational understanding, urban perception theory provides more granular insights into feature distinctiveness. Lynch's seminal work \citep{lynchImageCity2008} on urban legibility identifies landmarks—physical references distinguished by their singularity or memorable form—as critical anchors for spatial recognition, with hierarchical distinctiveness: an architecturally unique civic building possesses greater "landmark value" than a generic residential structure, even though both represent permanent infrastructure. Similarly, Golledge's \citep{golledge1999wayfinding} wayfinding behavior framework and Sorrows and Hirtle's \citep{sorrowsNatureLandmarksReal1999} landmark salience taxonomy, distinguish between perceptually salient elements that serve as distinctive reference points (e.g., ornate building details, unique signage with high visual or cognitive prominence) and background elements forming the spatial context (e.g., repetitive patterns, standard street furniture with low salience). This suggests that finer-grained differentiation based on spatial uniqueness within the broader urban environment—distinguishing common architectural patterns from rare, localizable features—can further enhance place recognition under adverse conditions.

Our framework operationalizes these principles by prompting the LLM to direct attention toward distinctive urban elements rather than generic scene components based on their spatial uniqueness within the broader cityscape context. Rather than simply segmenting objects, the LLM assesses their contribution to place distinctiveness—effectively distinguishing between common architectural patterns abundant citywide and rare, localizable features. LLMs, with their advanced multimodal reasoning capabilities, offer a novel mechanism to provide such context-aware spatial guidance, particularly when domain shifts degrade the reliability of learned visual representations. \change{For the complete prompt, please refer to Supplementary Material S3.1.1.}

A challenge, however, is that although LLMs can analyze images by extracting features from specific regions, they cannot inherently output explicit spatial attention positions. The task of guiding LLMs to generate coordinates with spatial information within images is known as visual grounding and recent research has explored various strategies for aligning LLM responses to pixel-wise regions within images. For instance, GLaMM \citep{rasheedGLaMMPixelGrounding2024} introduces a unified framework capable of generating natural language responses seamlessly intertwined with pixel-level segmentation masks, enabling grounded conversation generation. Similarly, VPP-LLaVA \citep{tangVisualPositionPrompt2025} leverages visual position prompts to enhance LLaVA’s training performance in visual grounding tasks by explicitly encoding spatial cues.

To operationalize this process, we propose an axis-based visual prompting strategy that positions coordinate axes outside the image frame to provide stable spatial references without altering visual content (Figure \ref{fig:method_attgen}). This design creates a lightweight and interpretable scaffold that enables the LLM's text-based reasoning to be spatially grounded during inference. The axis-based approach offers clear advantages over alternative visual prompting methods: grid-based overlays introduce visual artifacts that interfere with the SVIs, while bounding-box references, though providing accurate localization, require additional computation and exhibit failure cases in complex scenes. Consistent with insights from prior multimodal grounding research, including visual position prompting methods \citep{tangVisualPositionPrompt2025}, our approach benefits from explicit positional cues that strengthen the alignment between textual reasoning and visual localization. Qualitative inspection of the LLM-generated attention maps revealed strong alignment between the highlighted regions and the model's textual reasoning, indicating consistent spatial grounding. These spatial maps encapsulate intermediate spatial reasoning from the LLM and serve as inputs to the subsequent interpolation process for constructing the full attention map.

To bridge the gap between semantic reasoning and visual representation, we convert the discrete attention points provided by the LLM into a continuous spatial attention map. The LLM outputs a set of \change{normalized coordinates $c_i$ and associated importance weights $w_i$}. We employ Radial Basis Function (RBF) interpolation with a Gaussian kernel to construct a smooth attention surface $A \in \mathbb{R}^{H \times W}$ over the feature grid.

This process functions similarly to kernel density estimation in spatial analysis: locations explicitly identified by the LLM act as centers of influence, while the RBF kernel ensures a smooth transition of attention values across the image. The bandwidth parameter ($\sigma = 0.2$) was empirically selected to balance localization precision with smooth spatial blending. The resulting map is normalized and clamped to the range $[0, 2]$, ensuring that the attention mechanism can both up-weight discriminative features and suppress noise without causing numerical instability in the subsequent network layers.

\subsection{Attention Integration into VPR models for Inference}
\label{sec:att_model}

\begin{figure}[pos=htbp] 
    \centering
    \includegraphics[width=1\textwidth]{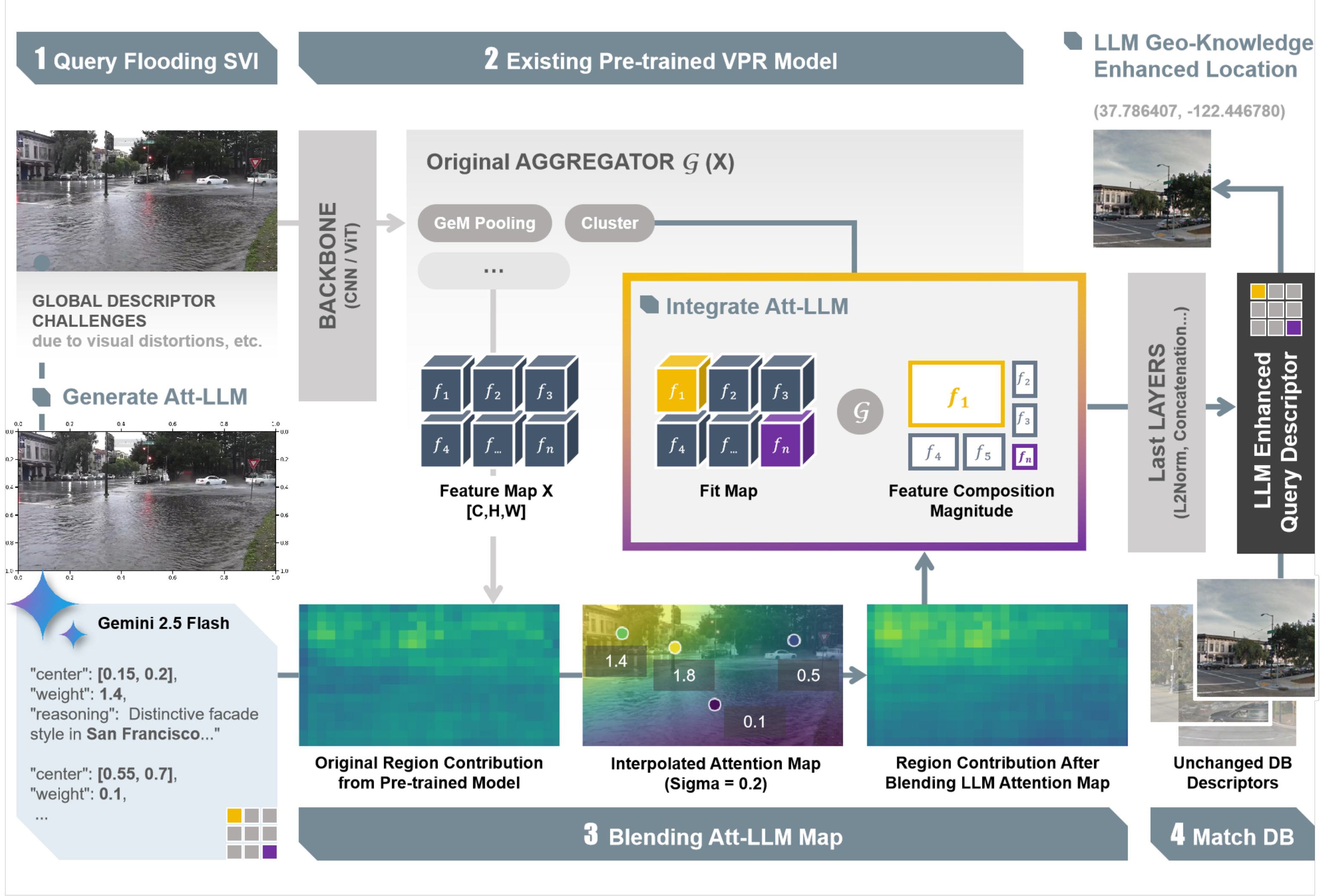} 
    \caption{\textbf{Overview of the VPR-AttLLM Framework for Enhanced Global Descriptor in Visual Place Recognition.} The framework integrates large language model (LLM) attention guidance into existing pre-trained VPR models through a plug-and-play modification of the aggregator module. The pipeline comprises four stages: \textbf{(1) Query Flooding with LLM Attention Generation}: A query street-view image depicting flooding conditions is processed by an LLM (\emph{e.g.}, Gemini-2.5-Flash) with geo-contextual prompts to generate spatial attention weights emphasizing scene elements distinctive for localization in the specified urban context (\emph{e.g.}, San Francisco). \textbf{(2) Existing Pre-trained VPR Model}: Feature extraction using established VPR architectures with diverse backbones (\emph{e.g.}, VGG16, ResNet50, DINOv2) produces a feature map $\mathbf{X} \in \mathbb{R}^{C \times H \times W}$. \textbf{(3) Blending LLM Attention Map}: The aggregator module integrates the LLM attention: (left) original spatial contribution map from the pre-trained model, (center) interpolated LLM attention map ($\sigma = 0.2$) highlighting architecturally stable elements, (right) blended contribution map demonstrating enhanced focus on distinctive, localization-relevant structures. \textbf{(4) Database Matching}: The LLM-enhanced query descriptor is matched against unchanged database descriptors from the original VPR model using cosine similarity, enabling seamless integration with existing databases without reprocessing.
} 
    \label{fig:method_model}  
\end{figure}

We integrate the LLM-generated attention maps into pre-trained VPR models using a modulation strategy that preserves the backbone's learned structure while injecting semantic guidance. The overall VPR-AttLLM framework is illustrated in Figure~\ref{fig:method_model}. Critically, this integration is \textit{asymmetric}: it applies only to query images at inference time. This allows the system to adapt to challenging scenes (e.g., floods, occlusions) without the prohibitive computational cost of re-processing the entire reference database.

\subsubsection{General Framework Formulation}

Let $\mathcal{G}$ be the aggregation module of a pre-trained VPR model that transforms a spatial feature map $\mathbf{X}$ into a global descriptor vector \change{$\mathbf{V}$}. We introduce the LLM-generated attention map $\mathbf{A}^{\text{LLM}}$ into this aggregation process, controlled by a blending coefficient $\alpha \in [0, 1]$. The attention-modulated inference is defined as:

\begin{equation}
\mathbf{V}^{\text{LLM}} = \mathcal{G}(\mathbf{X}; \mathbf{A}^{\text{LLM}}, \alpha)
\end{equation}

Here, $\alpha$ acts as a hyperparameter governing the influence of semantic reasoning. When $\alpha = 0$, the standard pre-trained behavior is preserved; as $\alpha$ approaches 1, the descriptor formation is increasingly dominated by the LLM's semantic priors. This formulation allows us to emphasize discriminative regions (e.g., permanent landmarks) and suppress transient noise by scaling the contribution of specific spatial locations in the feature map.

\subsubsection{Modulation for GeM Pooling and Cluster Aggregation}
\label{method_variants}

We adapt this general principle to two dominant aggregation paradigms: Generalized Mean (GeM) pooling and Cluster-based aggregation.

\textbf{GeM Pooling Integration.} Standard GeM pooling aggregates features by computing a spatially weighted average based on activation intensity. We modify this by blending the model's native spatial weights $\mathbf{W}^{\text{GeM}}$ with our semantic attention map $\mathbf{A}^{\text{LLM}}$. The final weight map is defined as $\mathbf{W}^{\text{final}} = \mathbf{W}^{\text{GeM}} + \alpha (\mathbf{A}^{\text{LLM}} - \mathbf{W}^{\text{GeM}})$. Consequently, the attention-weighted GeM descriptor is computed as:

\begin{equation}
\mathbf V_c^{\text{LLM}} = \left( \frac{\sum_{i,j} \mathbf{W}^{\text{final}}_{i,j} \cdot x_{c,i,j}^p}{\sum_{i,j} \mathbf{W}^{\text{final}}_{i,j}} \right)^{1/p}
\end{equation}

where $x_{c,i,j}$ represents the feature activation and $p$ is the pooling parameter. This ensures that regions deemed semantically important by the LLM contribute more significantly, correcting cases where the visual model focuses on salient but irrelevant or less identifiable features within given urban context.

\textbf{Cluster-based Integration.} For architectures like NetVLAD or SALAD, which assign local features to semantic clusters before aggregation, we utilize the attention map to modulate feature magnitude. Rather than altering the cluster assignment probabilities—which represent the visual "type" of the feature—we scale the feature vector $\mathbf{F}_{:,j}$ by its corresponding attention value $A^{\text{LLM}}_j$ prior to aggregation:

\begin{equation}
\mathbf{V^{\text{LLM}}}_{:,k} = \sum_{j=1}^{n} P_{k,j} \cdot \left( \alpha \cdot A^{\text{LLM}}_j \cdot \mathbf{F}_{:,j} \right)
\end{equation}

where $P_{k,j}$ is the assignment probability of spatial unit $j$ to cluster $k$. This acts as a magnitude modulator: a unique building (high attention) will exert a stronger influence on the "building" cluster representation than a generic wall (low attention), enhancing the discriminative power of the resulting global descriptor.

\section{Experiments} 
\label{sec:experiments}

\subsection{Datasets}
\label{sec:datasets}

To emulate real-world social media-driven geo-localization scenarios, our evaluation leverages city-wide reference databases that approximate the full urban extent of practical VPR deployment. This configuration enables assessment of LLMs' broad geographic reasoning capabilities under realistic retrieval conditions.

\begin{figure}[pos=htbp]
    \centering
    \includegraphics[width=\textwidth]{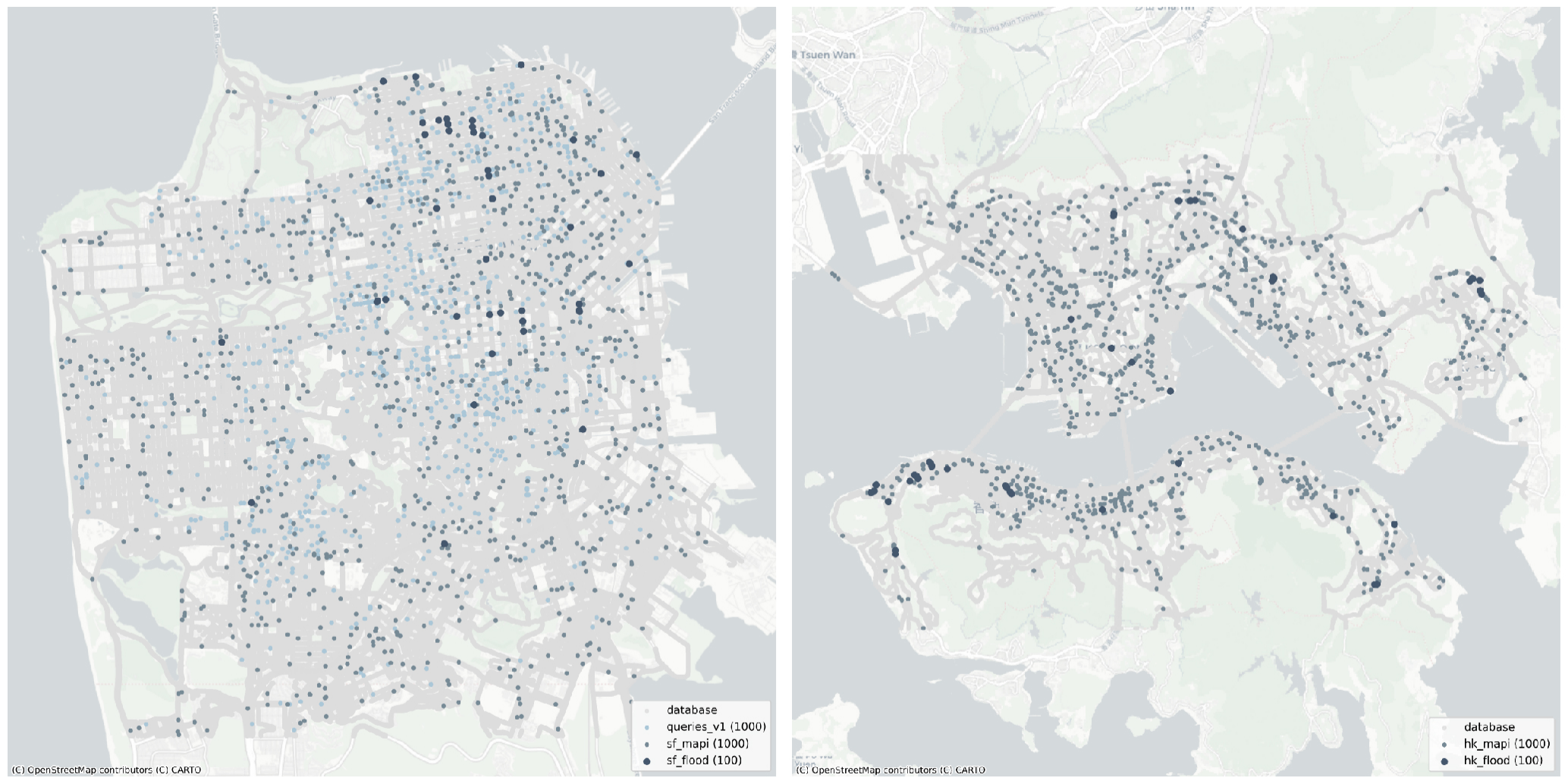} 
    \caption{Geographic distribution of experimental datasets. \textbf{(Left)} SF-XL dataset covering San Francisco: light gray dots indicate reference database locations ($\sim$2.8M images), light blue shows original benchmark queries (\texttt{sf\_v1}, 1000 images), medium blue represents sampled Mapillary images (\texttt{sf\_mapi}, 1000 images), and dark blue denotes real flooding images from social media (\texttt{sf\_flood}, 100 images). \textbf{(Right)} HK-URBAN dataset covering Hong Kong: reference database contains $\sim$1.15M images from 95,760 locations, with 100 real social media flooding images (\texttt{hk\_flood}), 1000 sampled Mapillary images (\texttt{hk\_mapi}), and 1000 synthetically generated flooding scenarios (\texttt{hk\_mapi[F]}).}
    \label{fig:experiment_map}  
\end{figure}

\subsubsection{SF-XL Dataset and Query Construction}
Among existing VPR benchmarks (e.g., Pitts250k~\citep{torii247Place2018}, Nordland, SF-XL~\citep{bertonRethinkingVisualGeolocalization2022a}), only SF-XL provides urban-scale ground-truth coverage suitable for city-wide evaluation. The SF-XL reference database contains 2.81 million geo-referenced images sampled from 232k unique locations across San Francisco, making it an ideal testbed for assessing VPR performance at metropolitan scale.

Building upon SF-XL, we construct challenging flood-related query sets that prioritize real-world validation while enabling controlled evaluation. Our evaluation strategy consists of three complementary components: (1)~\textbf{Real crowdsourced flooding images}: 100 geo-tagged flood photographs manually collected from social media platforms across San Francisco (\texttt{sf\_flood}), providing authentic ground truth for extreme weather conditions; (2)~\textbf{Controlled synthetic flooding on benchmark data}: Generative augmentation applied to 1,000 original SF-XL queries (\texttt{sf\_v1[F]}) to create systematic flooding variations while maintaining exact spatial correspondence with database images; and (3)~\textbf{Contemporary street-view augmentation}: 1,000 recent Mapillary images (\texttt{sf\_mapi}) with corresponding simulated flood scenes (\texttt{sf\_mapi[F]}), bridging temporal gaps in benchmark databases. This multi-faceted approach enables both authentic real-world validation and controlled studies under precisely known correspondences.

\begin{figure}[pos=htbp] 
    \centering
    \includegraphics[width=\textwidth]{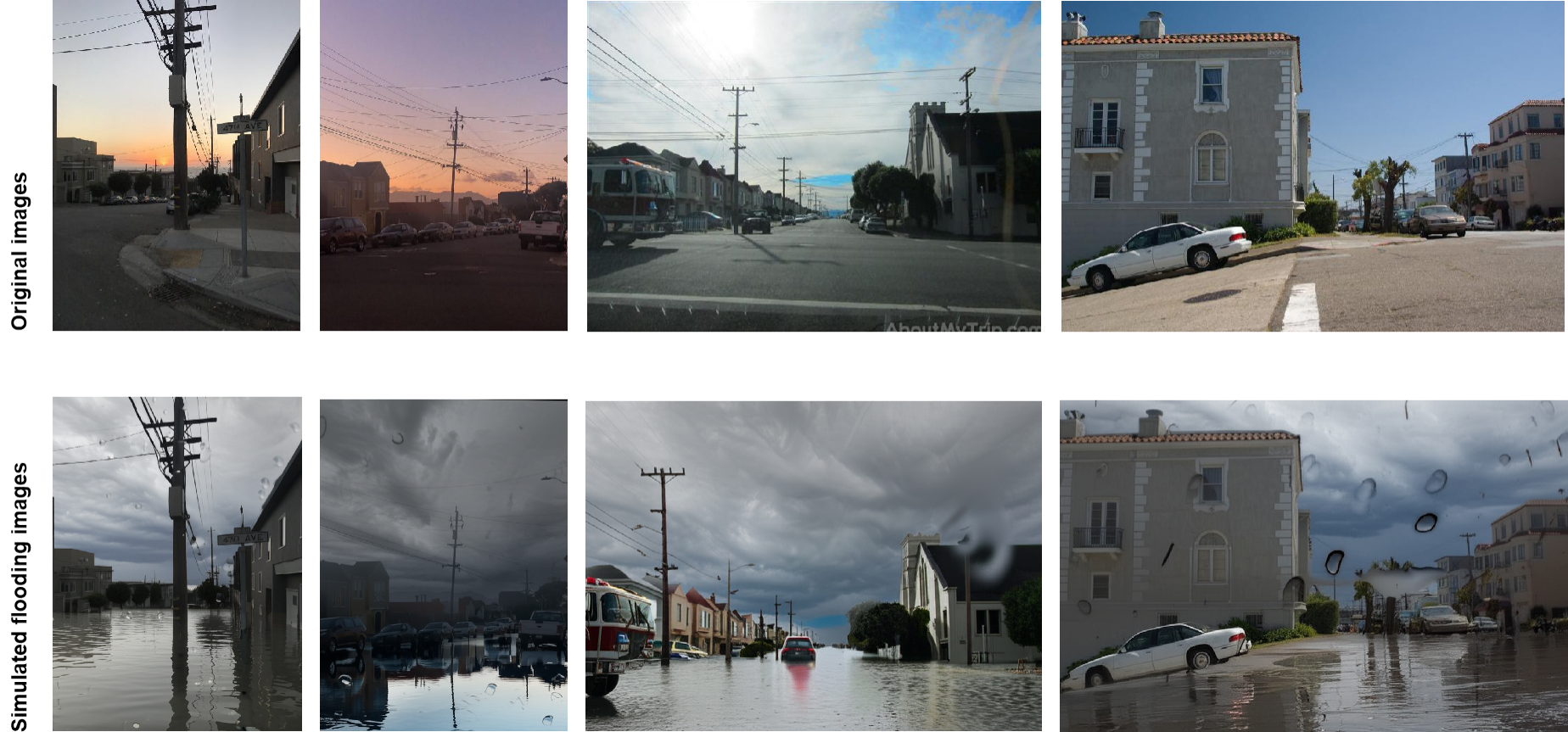} 
    \caption{Examples of simulated flooding applied to original query images, demonstrating controlled degradation while preserving geometric structure and viewpoint.} 
    \label{fig:experiment_sim}
\end{figure}

To ensure realism while maintaining geometric consistency essential for VPR evaluation, we employ a carefully designed prompting protocol (detailed in Supplementary Material Section S3.2. Prompts used for simulating flooding queries) using Gemini-2.0-Flash. Following best practices in synthetic data generation for SVI vision tasks~\citep{tremblayRainRenderingEvaluating2021, poravDontWorryWeather2019, torii247Place2018}, our prompts explicitly instruct preservation of original viewpoints and spatial layouts while introducing only appearance degradations: partial inundation, raindrop blur, surface reflections, and atmospheric haze (Figure~\ref{fig:experiment_sim}). \textbf{Critically, all generated images undergo manual verification to confirm that viewpoints and permanent structural elements remain unmodified}—only extreme weather effects are added. This view-preserving protocol ensures that any performance differences stem purely from appearance variations rather than geometric inconsistencies, enabling rigorous controlled evaluation of condition-invariant VPR methods.

\begin{table}[pos=htbp][htbp]
    \centering
    \begin{tabularx}{\textwidth}{p{0.17\textwidth}p{0.15\textwidth}p{0.06\textwidth}X}
        \toprule
        \textbf{Reference Database} & \textbf{Query Dataset} & \textbf{Size} & \textbf{Description} \\
        \midrule
        \multirow{5}{=}{\textbf{SF-XL}\\ 2.81M images\\ 232K locations} 
        & \texttt{sf\_flood} & 100 & Real flooding images from social media platforms \\
        & \texttt{sf\_v1} & 1000 & Original benchmark queries \\        
        & \texttt{sf\_v1[F]} & 1000 & Synthetic flooding applied to benchmark queries \\
        & \texttt{sf\_mapi} & 1000 & Mapillary images sampled across database coverage \\        
        & \texttt{sf\_mapi[F]} & 1000 & Synthetic flooding applied to Mapillary samples \\
        \midrule
        \multirow{3}{=}{\textbf{HK-URBAN}\\ 1.15M images\\ 95.8K locations} 
        & \texttt{hk\_flood} & 100 & Real flooding images from social media platforms \\
        & \texttt{hk\_mapi} & 1000 & Mapillary images sampled across database coverage \\
        & \texttt{hk\_mapi[F]} & 1000 & Synthetic flooding applied to Mapillary samples \\
        \bottomrule
    \end{tabularx}
    \caption{Summary of SF-XL and HK-URBAN datasets used in experiments.}
    \label{tab:experiment_dataset}
\end{table}

\subsubsection{HK-URBAN: A New Dataset with Distinctive Urban Morphology}
To complement SF-XL evaluation and examine performance in a distinct urban morphology, we construct \textbf{HK-URBAN}, a new dataset covering Hong Kong's primary urban areas with spatial scope comparable to SF-XL. The reference database comprises approximately 1.15 million perspective images extracted from 95,760 geographic locations spanning both Kowloon and Hong Kong Island.

Database construction follows established VPR protocols~\citep{torii247Place2018, bertonRethinkingVisualGeolocalization2022a}. Specifically, Google Street View panoramas are sampled at 15\,m intervals along all accessible streets within Hong Kong's central districts. From each panorama, twelve perspective views are rendered with $60^{\circ}$ horizontal field of view (FOV) and $12^{\circ}$ upward pitch to capture building facades and skyline features critical for urban VPR. Query sets mirror those constructed for SF-XL: 100 real social media flood images (\texttt{hk\_flood}), 1,000 Mapillary samples (\texttt{hk\_mapi}), and 1,000 corresponding simulated flood scenarios (\texttt{hk\_mapi[F]}).

HK-URBAN extends evaluation beyond Western urban morphology, enabling assessment of VPR robustness in a dense, high-rise Asian metropolis characterized by distinct architectural styles, vertical urban development, and complex street-level environments. The geographic distribution of all databases and query sets is illustrated in Figure~\ref{fig:experiment_map}, with dataset specifications summarized in Table~\ref{tab:experiment_dataset}.

\subsection{Implementation Details}
\label{sec:implementation}

To demonstrate the generalizability of the proposed VPR-AttLLM framework, we conduct extensive experiments across diverse state-of-the-art VPR architectures with different backbone networks and aggregation mechanisms. For CNN-based paradigms employing Generalized Mean (GeM) pooling as the final aggregator, we evaluate CosPlace~\citep{bertonRethinkingVisualGeolocalization2022a} and EigenPlaces~\citep{bertonEigenPlacesTrainingViewpoint2023} using their publicly released 512-dimensional descriptors with VGG16 and ResNet backbones, respectively. For Vision Transformer (ViT)-based paradigms with cluster-based aggregation, we evaluate SALAD~\citep{izquierdoOptimalTransportAggregation2024} in both its full configuration (8192D patch features $+$ 256D class token) and compact variant (512D patch features $+$ 32D class token), where the first component represents spatially aggregated patch-level descriptors and the second captures global scene semantics from the ViT class token.

To contextualize the effectiveness of our training-free, post-processing framework, we compare against Query Expansion (QE), a widely adopted post-retrieval refinement technique in VPR. In our experiments, we implement average QE with weighting parameter $\alpha = 0.8$ for the original query descriptor and $(1-\alpha)$ distributed equally among the top-$k$ retrieved reference descriptors, where $k \in \{3, 5\}$. This configuration is tested across all three pre-trained models (CosPlace, EigenPlaces, and SALAD) to provide fair comparison against VPR-AttLLM under identical retrieval conditions.

All experiments utilize officially released pre-trained weights without fine-tuning to ensure reproducibility and fair comparison. Backbone networks remain frozen throughout evaluation, with LLM-generated attention maps integrated solely through modified aggregation layers (either LLM-guided GeM pooling or cluster aggregation, as detailed in Section~\ref{sec:att_model}). Spatial attention maps for all query images are generated using Gemini-2.5-Flash, selected for its extensive geographic knowledge, robust multimodal reasoning capabilities, and particularly strong performance in spatial grounding tasks. The attention map generation method is detailed in Section~\ref{sec:att_gen}.

Following standard evaluation protocols in visual place recognition, performance is measured using Recall@N (with $N = 1, 5, 10$), which denotes the percentage of queries whose top-$N$ retrieved images are within a spatial threshold of the ground-truth location. For the \texttt{sf\_flood} and \texttt{hk\_flood} query subsets—curated from social media and subject to limited geolocation precision—a 100\,m threshold is adopted to accommodate positional noise. For all other queries derived from benchmark or Mapillary sources with accurate geotags, a 25\,m threshold is used, aligning with standard practice in VPR evaluations.

\subsection{Results}
\label{sec:results}

\added{This section presents the quantitative evaluation of the proposed VPR-AttLLM framework. We first establish its retrieval performance under severe real and synthetic flooding conditions (Section~\ref{sec:result_challenge}), followed by an assessment of its baseline stability in uncorrupted, common scenarios (Section~\ref{sec:result_common}). Finally, to validate the operational robustness and reproducibility of the framework, we present a prompt sensitivity analysis (Section\ref{sec:result_prompt}) and a cross-model evaluation utilizing open-source LLM alternatives (Section~\ref{sec:result_llm}).}

\subsubsection{Performance Under Flooding Scenarios}
\label{sec:result_challenge}

\begin{figure}[pos=htbp]
    \centering
    \includegraphics[width=\textwidth]{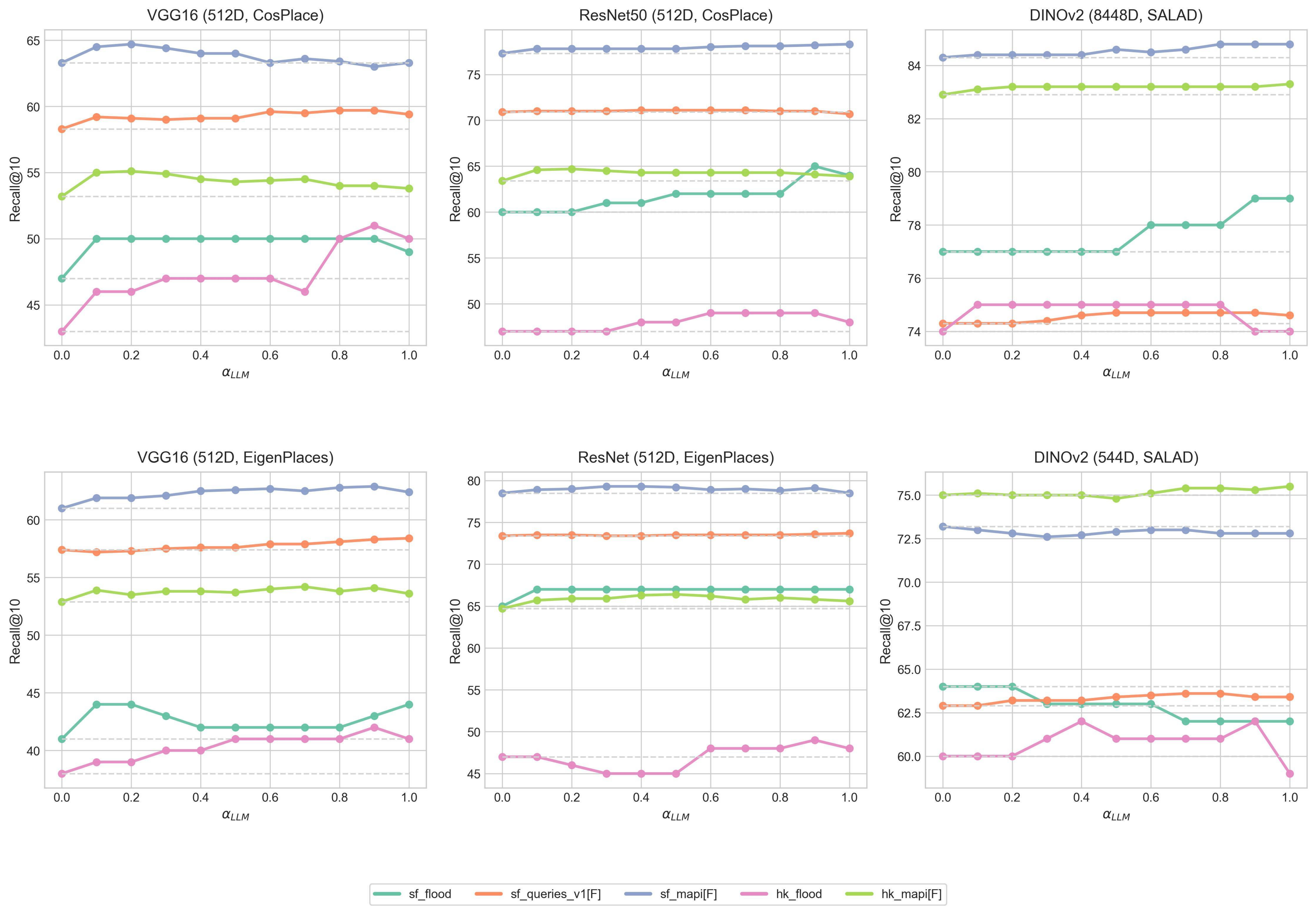} 
    \caption{Recall@10 performance as a function of LLM attention weight $\alpha_{\text{LLM}}$ across six VPR architectures and five flooding query sets. Each subplot shows results for a specific model-backbone combination: CosPlace and EigenPlaces with VGG16 (512D) and ResNet50 (512D) backbones, and SALAD with DINOv2 backbone in both full (8448D) and compact (544D) configurations. The baseline performance ($\alpha_{\text{LLM}} = 0.0$) is shown as the leftmost point. VPR-AttLLM consistently improves Recall@10 across nearly all conditions, with real-world flooding scenarios (\texttt{hk\_flood}, \texttt{sf\_flood}) exhibiting the most substantial gains. Performance remains stable or improving across the full range $\alpha_{\text{LLM}} \in [0.1, 1.0]$, with only isolated minor degradation observed at $\alpha_{\text{LLM}} = 1.0$ in limited test cases (e.g., \texttt{hk\_flood}).}
    \label{fig:experiment_challenging}  
\end{figure}

\paragraph{\change{Consistent Improvement and Parameter Stability}}
\change{Under both real and synthetic flooding conditions, the LLM-guided attention mechanism (VPR-AttLLM) demonstrates consistent improvements over unenhanced baseline models ($\alpha_\text{LLM}=0.0$). Figure}~\ref{fig:experiment_challenging} \change{illustrates the Recall@10 variations across all query sets as the attention weight $\alpha_\text{LLM}$ increases. As shown, VPR-AttLLM consistently elevates recall rates above the baseline across nearly the entire tested range of $\alpha_\text{LLM} \in [0.1,1.0]$. The method exhibits high parameter stability, with majority $\alpha_\text{LLM}$ reliably optimizing semantic-visual integration. The most pronounced gains occur in real-world flood imagery (\texttt{sf\_flood}, \texttt{hk\_flood}), where conventional visual features suffer the most severe degradation. In contrast, standard Query Expansion (QE) applied across the same backbones either degraded performance or yielded marginal improvements (Table S1), confirming VPR-AttLLM as a superior enhancement strategy.}

\paragraph{\change{Robustness Across Architectures}}
\change{The benefits of semantic attention modulation hold across varying backbone configurations. As seen in Figure}~\ref{fig:experiment_challenging}, \change{VGG16-based models achieved performance improvements exceeding 3\% across real flooding images. Most notably, CosPlace (VGG16) achieved an 8\% improvement in Recall@10 on the \texttt{hk\_flood} subset (from 43.0\% to 51.0\% at $\alpha_\text{LLM}=0.8$). This gain is highly significant given the model's lack of prior exposure to Hong Kong's dense high-rise morphology, indicating that LLM-derived semantic priors can effectively recover scene structure under severe domain shift. Similar robustness was observed in ResNet50 models, which peaked at a 5\% absolute gain on \texttt{sf\_flood}. Furthermore, explicit semantic guidance proved valuable even for advanced transformer models; the SALAD model (DINOv2) maintained steady gains of 0.5–1.0\% despite its already rich cross-modal pre-training.}

\begin{figure}[pos=htbp]  
    \centering
    \includegraphics[width=0.5\textwidth]{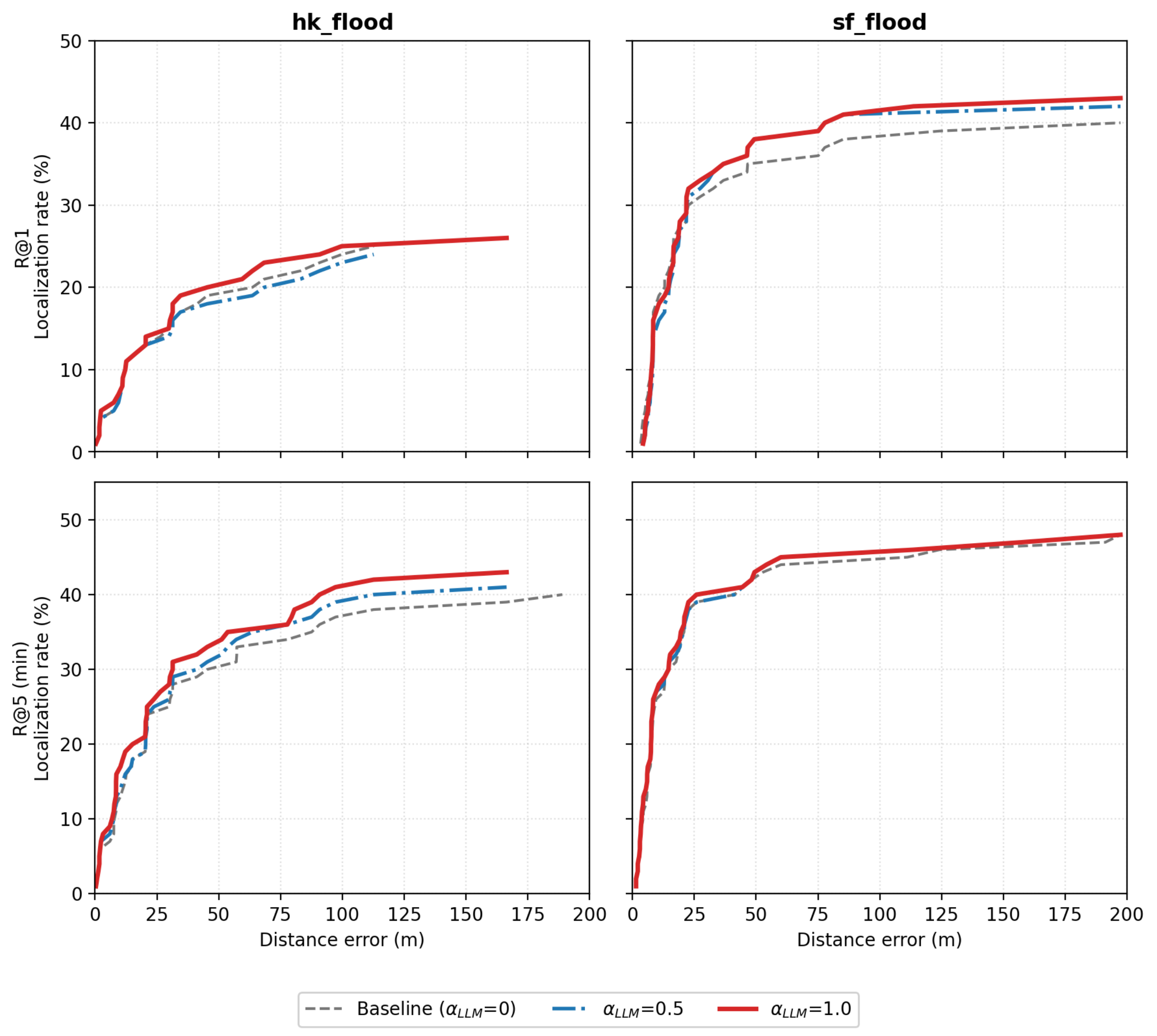} 
    \caption{\added{CDF of localization distance errors for datasets hk\_flood and sf\_flood under R@1 (top) and R@5-min (bottom) metrics using CosPlace (VGG16). Higher $\alpha_{\text{LLM}}$ values consistently shift curves leftward, indicating improved localization accuracy within tight spatial thresholds relative to the unenhanced baseline.}}
    \label{fig:experiment_cdf_cosplace}  
\end{figure}

\paragraph{\added{Geolocalization Accuracy and Operational Utility}}
\added{While Recall@K evaluates the retrieval of exact reference matches within a predefined radius, practical disaster management requires minimizing the absolute physical distance error of geolocalization. To evaluate operational utility, Figure}~\ref{fig:experiment_cdf_cosplace} \added{presents the Cumulative Distribution Function (CDF) of the localization distance error (in meters) for the two real-world flood datasets (hk\_flood and sf\_flood), demonstrated using CosPlace (VGG16) as baseline. Extra CDF analysis plot can be found in Figure S1}

\added{As illustrated in this figure, the distributions for R@1 and the minimum distance within R@5 demonstrate that the LLM-guided attention mechanism consistently shifts a higher percentage of retrieved images to tighter distance thresholds. For instance, at $\alpha_{\text{LLM}} \in {0.5, 1.0}$, the framework successfully localizes a higher volume of queries within the critical 0–100 meter range compared to the unenhanced baseline $\alpha_{\text{LLM}} = 0.0$. In real-world flooding scenarios, this reduction in absolute spatial error translates to more precise pinpointing of critical infrastructure damage from crowdsourced imagery, directly enhancing the utility of VPR systems for emergency responders.}

\subsubsection{Performance Under Common Scenarios}
\label{sec:result_common}

\change{To ensure the LLM-guided attention mechanism does not compromise baseline performance during routine operation ("do no harm"), we evaluated the framework on three query sets lacking severe appearance degradation: \texttt{sf\_v1}, \texttt{sf\_mapi}, and \texttt{hk\_mapi} (Table S2). As expected, the effects of $\alpha_{\text{LLM}}$ under these common scenarios are considerably more subtle. On the benchmark \texttt{sf\_v1} set, where visual patterns are clearly defined and closely align with the reference database, performance changes are marginal. This reflects the limited necessity of external semantic guidance when baseline visual descriptors are already highly discriminative. However, on the more heterogeneous Mapillary-based queries (\texttt{sf\_mapi} and \texttt{hk\_mapi}), the framework yields modest but consistent improvements (approximately 1\% in Recall@1 and Recall@5) for both CosPlace and EigenPlaces backbones. This indicates that VPR-AttLLM provides incremental stability even under subtle domain shifts caused by varying camera viewpoints, capture devices, or temporal lighting changes.}

\change{Crucially, this baseline stability highlights a major architectural advantage over traditional Query Expansion (QE). Across all tested configurations (k=3 and k=5), QE consistently degraded baseline performance on these common sets. Because QE relies entirely on initial retrieval quality, geographically distant false-positive candidates corrupt the expanded query descriptor and compound retrieval errors. In contrast, VPR-AttLLM operates independently of initial retrieval outcomes by applying semantic attention modulation directly at the feature aggregation stage. This proactive design ensures that the system safely defaults to baseline visual matching under ideal conditions, while remaining robust enough to provide semantic rescue when visual cues are disrupted.}

\subsubsection{\added{Impact of Prompt Sensitivity}}
\label{sec:result_prompt}

\added{To understand how instruction complexity impacts the LLM-guided attention mechanism, we conducted a rigorous prompt sensitivity analysis. We evaluated our primary model (Gemini-2.5-Flash) across three prompt variations: the original "Full structured prompt" (which includes complex routing and an explicit 'None' output option), a "City name switched" prompt (to test spatial bias), and a "Minimal prompt" designed to test the lower bounds of instruction requirements.}

\newcolumntype{P}{>{\centering\arraybackslash}p{0.075\textwidth}}
\newcolumntype{M}{>{\raggedright\arraybackslash}p{0.2\textwidth}}

\begin{table}[pos=htbp][htbp]
\centering
\small
\begin{tabularx}{\textwidth}{X M P P P P P P}
\toprule
\multirow{2}{*}{\textbf{Base Model}} & \multirow{2}{*}{\textbf{Prompt type}} & \multicolumn{3}{c}{\textbf{sf\_flood}} & \multicolumn{3}{c}{\textbf{hk\_flood}} \\
\cmidrule(lr){3-5} \cmidrule(lr){6-8}
& & @1 & @5 & @10 & @1 & @5 & @10 \\
\midrule
\multirow{4}{=}{\textbf{VGG16 (512D, EigenPlaces)}} & \textbf{Baseline} & 38 & 40 & 41 & 18 & 33 & 38 \\
& Full structured & 39 & 42 & 44 & 19 & 36 & 41 \\
& City name switched & 36 & 41 & 42 & 20 & 37 & 42 \\
& Minimal & 40 & 44 & 45 & 19 & 34 & 39 \\
\midrule
\multirow{4}{=}{\textbf{ResNet50(512D, CosPlace)}} & \textbf{Baseline} & 52 & 57 & 60 & 30 & 45 & 47 \\
& Full structured & 53 & 60 & 64 & 31 & 42 & 48 \\
& City name switched & 53 & 59 & 64 & 35 & 44 & 51 \\
& Minimal & 53 & 59 & 62 & 31 & 44 & 51 \\
\bottomrule
\end{tabularx}
\caption{\added{Performance Comparison of different prompt types on flooding datasets.}}
\label{tab:comparison_prompt}
\end{table}

\added{As shown in Table~\ref{tab:comparison_prompt}, the system demonstrates high robustness to prompt phrasing. Surprisingly, reducing the instruction complexity to a "Minimal prompt" did not harm performance; rather, recall metrics remained highly comparable to the full structured prompt and occasionally yielded slight improvements. For instance, using the ResNet50 (CosPlace) backbone on the hk\_flood dataset, the Minimal prompt achieved a Recall@10 of 51\%, outperforming the Full structured prompt's 48\%. This indicates that the proposed attention-modulation framework relies more on the LLM's inherent visual-semantic grounding than on exhaustive prompt engineering.}

\subsubsection{\added{Robustness Across Open-Source LLMs}}
\label{sec:result_llm}

\added{To ensure the reproducibility of our method and reduce reliance on proprietary APIs, we evaluated the framework using alternative Vision-Language Models. We integrated Qwen3.5-Plus (accessed via API) and Qwen3-VL-8B-Instruct (an open-source model with locally deployable weights). During pilot testing, we observed that the smaller 8B model struggled with the complex logic of our original Full structured prompt, frequently exhibiting degenerate behavior by exploiting the 'None' exit condition. Consequently, we utilized the standardized "Minimal prompt" (validated in Section 4.3.2) for a fair comparison across all three models.}

\begin{table}[pos=htbp][htbp]
\centering
\small
\begin{tabularx}{\textwidth}{X M P P P P P P}
\toprule
\multirow{2}{*}{\textbf{Base Model}} & \multirow{2}{*}{\textbf{LLM}} & \multicolumn{3}{c}{\textbf{sf\_flood}} & \multicolumn{3}{c}{\textbf{hk\_flood}} \\
\cmidrule(lr){3-5} \cmidrule(lr){6-8}
& & @1 & @5 & @10 & @1 & @5 & @10 \\
\midrule
\multirow{4}{=}{\textbf{VGG16 (512D, EigenPlaces)}} & \textbf{Baseline} & 38 & 40 & 41 & 18 & 33 & 38 \\
& Gemini-2.5-flash & 40 & 44 & 45 & 19 & 34 & 39 \\
& Qwen3.5-plus & 38 & 42 & 42 & 20 & 35 & 39 \\
& Qwen3-vl-8b & 37 & 40 & 42 & 21 & 34 & 41 \\
\midrule
\multirow{4}{=}{\textbf{ResNet50 (512D, CosPlace)}} & \textbf{Baseline} & 52 & 57 & 60 & 30 & 45 & 47 \\
& Gemini-2.5-flash & 53 & 59 & 62 & 31 & 44 & 51 \\
& Qwen3.5-plus & 52 & 58 & 63 & 32 & 45 & 50 \\
& Qwen3-vl-8b & 52 & 58 & 61 & 31 & 46 & 47 \\
\bottomrule
\end{tabularx}
\caption{\added{Performance comparison of various LLMs integration on flooding datasets.}}
\label{tab:comparison_llm}
\end{table}

\added{The results (Table~\ref{tab:comparison_llm}) confirm that our attention-guidance architecture is model-agnostic. While the larger models generally exhibited the most stable gains due to superior spatial reasoning, the locally deployed Qwen3-VL-8B-Instruct model successfully improved upon the non-LLM baseline across both evaluated backbones. Notably, on the challenging hk\_flood dataset using the VGG16 (EigenPlaces) backbone, the 8B model even achieved a Recall@1 of 21\%, outperforming both Gemini-2.5-Flash (19\%) and the baseline (18\%), and Recall@10 of 41\%, outperforming the larger models' recall (39\%). To facilitate reproducibility outside of proprietary ecosystems, the implementation and deployment scripts for the open-source Qwen3-VL-8B-Instruct model are provided in the supplementary materials.}

\subsection{Interpretability Through LLM-Guided Attention}
\label{sec:interpretability}

Beyond quantitative improvements, the primary value of the VPR-AttLLM framework lies in its \textit{training-free} and \textit{highly interpretable} nature. Without any model fine-tuning or additional optimization, it provides measurable performance gains under extreme or visually unfamiliar conditions, demonstrating strong adaptability to the heterogeneous nature of real-world crowdsourced imagery. This section examines two representative cases—one involving viewpoint shift in common scenarios and another under extreme flooding conditions—to illustrate how LLM-derived semantic guidance complements conventional visual feature extraction.

\subsubsection{Qualitative Analysis of Success Cases}
\begin{figure}[pos=htbp]  
    \centering
    \includegraphics[width=\textwidth]{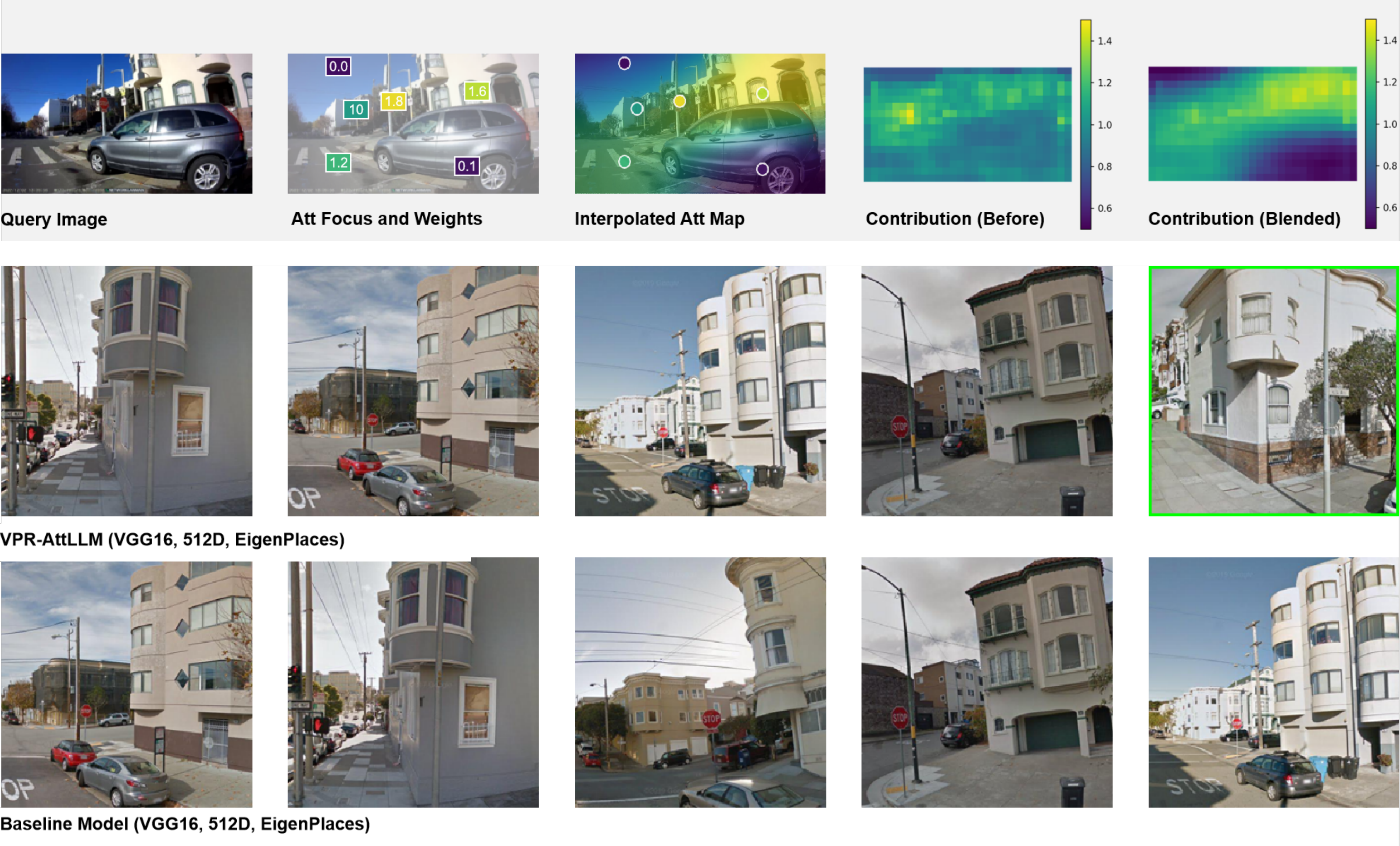} 
    \caption{
    Qualitative example of VPR-AttLLM in a common scenario with viewpoint shift (\texttt{sf\_mapi}).
    \textbf{Top row:} Query image, with blending LLM attention map's effects in spatial feature contribution. 
    \textbf{Medium row:} top-5 retrieval results from our VPR-AttLLM enhanced CosPlace, with the correct match highlighted in green.
    \textbf{Bottom row:} top-5 retrieval results from baseline CosPlace.
    Despite partial occlusion, the LLM identifies the right building's distinctive curved bay windows as a strong localization cue (weight 1.6), enabling successful retrieval within top-5 results by redirecting attention away from the fully visible but less architecturally unique left structure.
    }
    \label{fig:experiment_analysis1}  
\end{figure}

\paragraph{Case Study 1: Architectural Distinctiveness Under Viewpoint Shift} Figure~\ref{fig:experiment_analysis1} presents a successful retrieval case from the \texttt{sf\_mapi} test set, where the query exhibits moderate viewpoint shift. The original GeM pooling assigns higher feature contributions to the fully visible left building while down-weighting the partially occluded right structure, despite the latter's distinctive architectural characteristics. This spatial bias arises because conventional pooling mechanisms rely primarily on feature magnitude and spatial extent rather than semantic uniqueness. In contrast, the LLM automatically generates both a weight (1.6) and an interpretable justification for the right building: \textit{"Distinctive cream-colored building with curved bay windows and unique entrance, contributing significant architectural uniqueness."} By elevating attention to this architecturally unique element along with the adjacent street signage, VPR-AttLLM successfully retrieves the correct reference image within the top-5 results, demonstrating how semantic reasoning can override purely appearance-based feature weighting.

\begin{figure}[pos=htbp]  
    \centering
    \includegraphics[width=\textwidth]{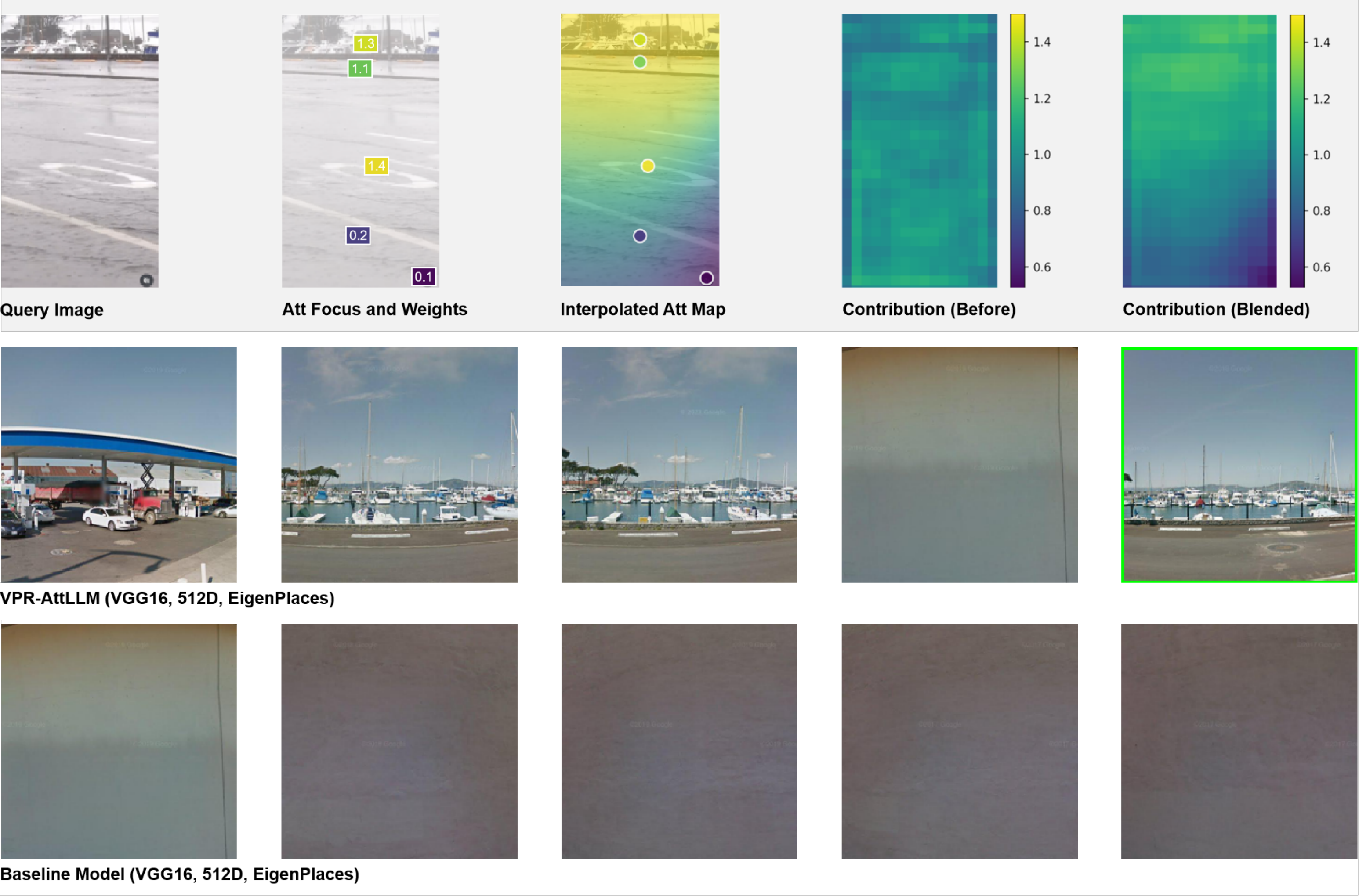} 
    \caption{
    Qualitative example of VPR-AttLLM under extreme flooding conditions (\texttt{sf\_flood}).
    \textbf{Top row:} Query image, with blending LLM attention map's effects in spatial feature contribution. 
    \textbf{Medium row:} top-5 retrieval results from our VPR-AttLLM enhanced EigenPlaces, with the correct match highlighted in green.
    \textbf{Bottom row:} top-5 retrieval results from baseline EigenPlaces.
    The baseline model, biased toward the flooded road surface, retrieves visually similar but semantically incorrect tunnel images. VPR-AttLLM identifies the upper marina landscape—boats, docks, and background infrastructure—as offering moderate to high localization cues (weight 1.3), successfully redirecting retrieval toward spatially proximate waterfront references.
    }
    \label{fig:experiment_analysis2}  
\end{figure}

\paragraph{Case Study 2: Stable Feature Selection Under Flooding} Figure~\ref{fig:experiment_analysis2} illustrates a more challenging case from the \texttt{sf\_flood} query set, where severe viewpoint tilt and focal distortion toward the flooded road surface compromise conventional feature extraction. The original GeM pooling fails to capture discriminative information from the upper portion of the image—a marina landscape with boats and docking infrastructure—resulting in retrieval of visually similar but semantically incorrect "gray" tunnel images from the reference database. Due to the scarcity of tilted viewpoint examples in the training data, the baseline model lacks robust priors for such extreme conditions. The LLM, however, readily identifies the upper region as semantically informative, automatically generating the reasoning: \textit{"The specific arrangement of boats and visible marina infrastructure (docks, background buildings, poles) offer moderate to high localization cues,"} and assigns a weight of 1.3 to this region. This semantic reweighting successfully redirects attention toward the marina scene, enabling retrieval of spatially proximate waterfront references and correcting the baseline's mislocalization.

\subsubsection{\added{Analysis of Failure Modes}}
\added{While LLM-guided attention generally improves geolocalization accuracy, understanding its limitations is essential for deploying these models reliably in complex urban environments. To provide a transparent evaluation, we conducted a qualitative analysis of instances where the introduction of LLM attention degraded recall performance compared to the baseline. As summarized in Table \ref{tab:failure_reason}, these failure cases typically originate from the complex interplay between the semantic prioritization of the LLM and the physical constraints of the reference Street View Imagery (SVI) database.}

\begin{table}[pos=htbp][htbp]
    \centering
    \begin{tabularx}{\textwidth}{p{0.18\textwidth}p{0.57\textwidth}X}
        \toprule
         \textbf{Failure Category}&  \textbf{Description}& \textbf{Primary Cause}\\
         \midrule
         Suboptimal Attention Allocation&  The LLM incorrectly prioritizes uninformative or transient foreground features (e.g., flooded street surfaces) over distinct structural landmarks.& LLM misinterpretation of scene context.\\
         \midrule
         Temporal Discrepancy &  The attention targets a valid landmark (e.g., a newly built bus stop), but the structure did not exist when the reference database SVI was captured.& Database obsolescence.\\
         \midrule
         Perceptual Aliasing on Large Structures&  The model correctly focuses on a massive landmark and retrieves it, but the matched image is from a different facade of the same building, exceeding distance thresholds.& Geometry of large-scale urban infrastructure.\\
         \midrule
         Database Viewpoint Deficiency&  The query features a highly specific perspective (e.g., an under-bridge view) that is entirely absent from the reference database, forcing a visually similar but incorrect match.& SVI database coverage limitations.\\
         \midrule
         Interpolation \& Resolution Artifacts&  The attention map correctly identifies an informative region, but the region is too small or heavily occluded by noise. Interpolating the attention map degrades the final aggregated descriptor.& Feature extraction granularity.\\
         \bottomrule
    \end{tabularx}
    \added{\caption{Summary of failure modes and their primary causes where LLM-guided attention degrades geo-localization performance.}}
    \label{tab:failure_reason}
\end{table}

\added{Our analysis indicates that performance bottlenecks fall into two broad categories: model-driven misattributions and database-driven constraints. On the model side, the LLM occasionally misinterprets scene context by prioritizing transient foreground elements, such as flooded surfaces, or the temporary billboard over permanent structural geometry. Furthermore, even when informative regions are correctly identified, insufficient feature extraction granularity can introduce resolution and interpolation artifacts if the target region is overly small or heavily occluded. Conversely, database-driven limitations reflect the inherent challenges of crowdsourced urban data. The dynamic nature of urban infrastructure frequently causes temporal discrepancies when queried landmarks postdate the reference SVI. Additionally, the massive scale of modern architecture introduces severe perceptual aliasing across identical building facades, while highly specific crowdsourced perspectives—such as under-bridge views—often suffer from viewpoint deficiencies within standard SVI coverage. A detailed visual breakdown of these representative failure modalities is provided in the Supplementary Materials (Figure S2).}

\subsubsection{Geographic Context in Semantic Attention}

To examine whether LLM-generated attention maps rely on city-specific semantic knowledge when evaluating architectural distinctiveness, we conduct a controlled ablation study by reversing the city context in attention generation prompts: specifying Hong Kong when processing San Francisco queries and vice versa. This experiment tests whether the same architectural element (e.g., a Victorian facade or a high-rise tower) receives different attention weights depending on its presumed urban context.

\begin{figure}[pos=htbp]
    \centering
    \includegraphics[width=0.7\textwidth]{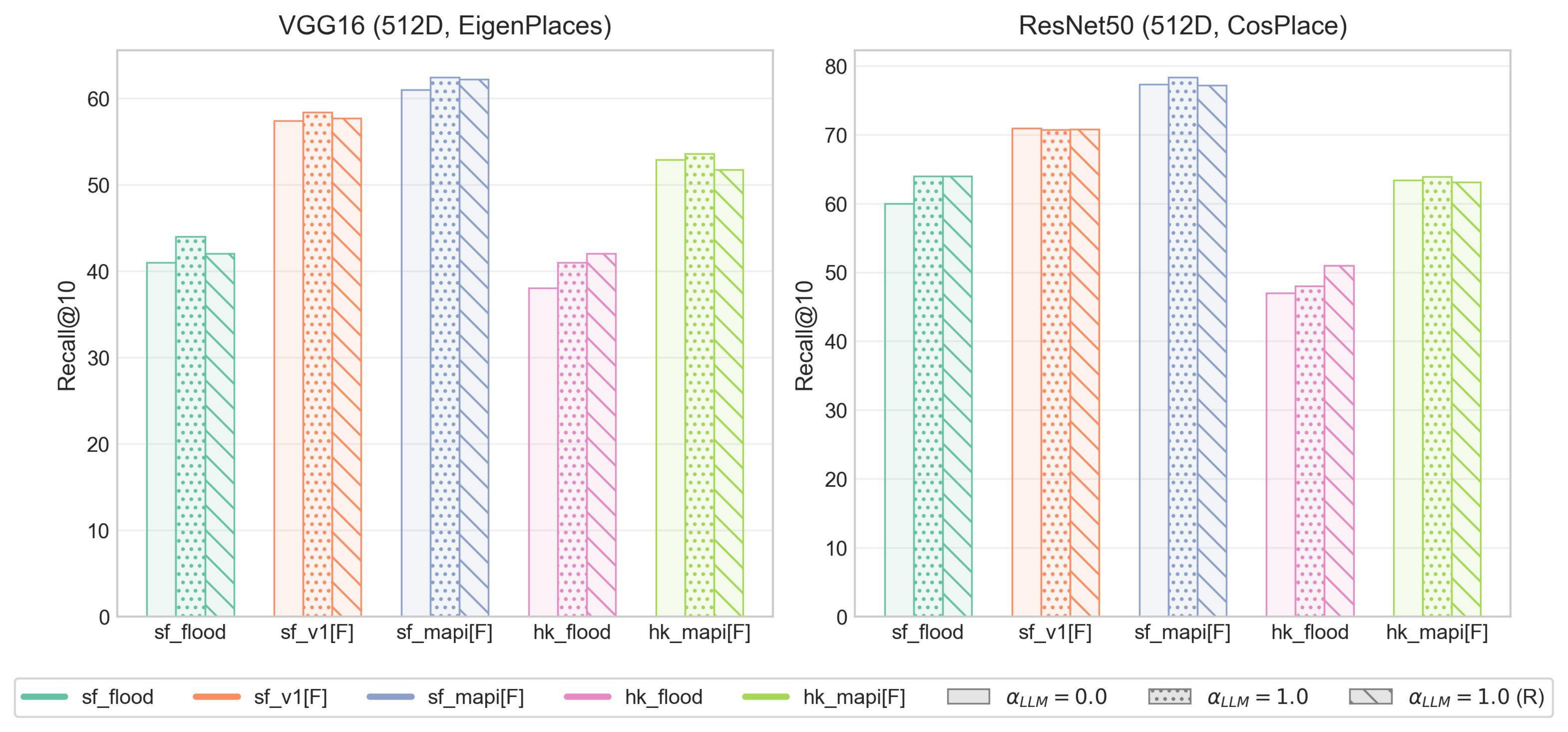}
    \caption{Recall@1, @5, and @10 comparison between correct and reversed city context across. Each subplot shows one model evaluated on all flooding query sets. For each metric, bars represent: baseline performance ($\alpha_{\text{LLM}}=0.0$, no hatching), correct city context at $\alpha_{\text{LLM}}=1.0$ (dot hatching) and reversed city context at $\alpha_{\text{LLM}}=1.0$ (diagonal hatching). Reversed prompts yield slightly lower performance than correct context but maintain substantial improvements over baseline under most cases, indicating robustness to imperfect geographic metadata.}
    \label{fig:reversed_prompt}
\end{figure}

Detailed quantitative results across all five flooding scenarios are provided in Supplementary Materials (Table S3). Figure~\ref{fig:reversed_prompt} presents recall results comparison, with reversed-prompt configurations denoted by "(R)" suffix. Reversed city context reduces Recall@10 by around 0.5\% compared to correct geographic grounding under most cases. For instance, on \texttt{sf\_mapi[F]}, CosPlace with ResNet50 achieves 78.3\% with correct context versus 77.2\% with reversed context ($\Delta = -1.1\%$); EigenPlaces with VGG16 shows 62.4\% versus 62.2\% ($\Delta = -0.2\%$)).

This modest but consistent degradation reveals two important interpretability insights. First, LLMs do encode city-specific geographic knowledge when assessing landmark distinctiveness. For example, single-family detached houses in Hong Kong are much rarer compared to San Francisco, constituting only a negligible proportion of housing stock limited to luxury enclaves, whereas they represent nearly one-third of San Francisco's residential units. Conversely, glass curtain-wall towers are more distinctive markers in SF than in HK's dense skyscraper clusters. This demonstrates that attention maps are not merely generic "interestingness" detectors but rather geographically contextualized evaluations of visual salience. Second, the relatively small magnitude of performance degradation (typically $<1.5\%$) and the fact that reversed-prompt configurations still outperform original baselines across most cases indicates the benefit derived from general semantic reasoning about architectural salience—identifying unique facades, landmark structures, and stable visual anchors—beyond precise city-specific knowledge. 

These findings carry important practical implications: the robustness of VPR-AttLLM to imperfect city context suggests that even without correct city contexts still provides sufficient general semantic grounding for meaningful improvements, lowering deployment barriers in scenarios where precise metadata may be unavailable or unreliable. More broadly, VPR-AttLLM serves as an effective \textit{plug-and-play semantic prior} that enhances both spatial reasoning and decision transparency without requiring retraining or domain-specific augmentation. The LLM-derived attention maps provide explicit semantic justifications—emphasizing distinctive urban features such as unique building facades, signage, and infrastructure elements while suppressing visually dominant but semantically uninformative regions (e.g., flooded streets, overexposed sky)—with particular value in challenging conditions where visual features alone prove insufficient. This interpretability through natural language reasoning enables practitioners to audit and refine model behavior in safety-critical applications, bridging the gap between high-performance deep learning and human-understandable decision-making in VPR systems.

\section{Discussion}  
\label{sec:discussion}

\change{While the preceding failure analysis highlights the inherent challenges of matching dynamic urban scenes to static reference databases, our overall findings demonstrate that training-free semantic guidance from large language models remains a robust tool for enhancing visual place recognition.} This is particularly evident when appearance degradation (e.g., flooding) compounds geographic domain shift. This section synthesizes key findings, examines the framework's limitations, and identifies opportunities for future integration of multimodal and geospatial reasoning in VPR systems.

\subsection{Effectiveness, model agnosticity, and interpretability of LLM-Guided attention}

The primary contribution of VPR-AttLLM lies in providing \textit{asymmetric query-side enhancement} that improves retrieval robustness without reprocessing reference databases—a critical advantage for deployment scenarios involving millions of indexed images. Across five challenging query sets, VPR-AttLLM consistently improved Recall@10 by 1--8\%, with the largest gains observed under two conditions: (1)~\textit{extreme flooding scenarios}, where CosPlace achieved up to 8\% improvement on real-world flood queries (\texttt{hk\_flood} and (2)~\textit{heterogeneous crowdsourced imagery}, where Mapillary queries (\texttt{sf\_mapi[F]}, \texttt{hk\_mapi[F]}) with diverse viewpoints and image quality benefited from 1--2\% recall improvements across both cities. Notably, performance gains on Hong Kong Mapillary queries (\texttt{hk\_mapi}, \texttt{hk\_mapi[F]}) were comparable to those on San Francisco (\texttt{sf\_mapi}, \texttt{sf\_mapi[F]}), despite models (CosPlace, EigenPlaces) being trained exclusively on San Francisco data. This suggests that LLM-guided attention enhances robustness to query heterogeneity in a cross-city transferable manner. The results validate VPR-AttLLM's dual capability in handling both extreme appearance degradation (flooding-induced texture distortion) and the intrinsic variability of real-world crowdsourced street-level imagery.

Beyond quantitative improvements, VPR-AttLLM addresses the undergeneralization problem in discriminative models \citep{ilievskiAligningGeneralizationHumans2025}, where VPR systems fail under extreme weather conditions such as flooding. The model-agnostic architecture enables seamless integration with diverse VPR backbones without retraining. The late-stage attention fusion strategy operates directly on feature aggregation \change{phase} for pre-trained global descriptors, making it compatible with both CNN-based models (CosPlace, EigenPlaces with VGG16 and ResNet50) and transformer-based architectures (Salad with DINOv2). This plug-and-play property facilitates rapid deployment and experimentation across existing VPR systems, as evidenced by consistent improvements across all three tested architectures. Furthermore, the framework enhances interpretability by providing natural language justifications for attention weights (e.g., "distinctive curved bay windows contributing significant architectural uniqueness"), allowing practitioners to audit which scene elements drive retrieval decisions—a critical requirement for safety-critical geospatial applications such as emergency response or autonomous navigation.

Finally, this work establishes a novel methodological bridge between urban perception theory and computer vision by operationalizing semantic place understanding through LLM-guided attention. The framework's ability to identify architecturally distinctive elements (e.g., unique facades, landmark structures) and suppress visually dominant but contextually uninformative regions (e.g., front large but common residential building) empirically validates urban theory about place uniqueness in a computational setting. By encoding human-like spatial reasoning—traditionally studied through cognitive mapping and environmental psychology—into feature weighting mechanisms, these interpretable attention maps generated by VPR-AttLLM offer potential value beyond VPR performance: they could serve as data-driven tools for validating urban design theories about place identity and wayfinding \citep{lynchImageCity2008}, or for auditing training dataset biases in visual geolocation systems. By making explicit which scene elements drive model decisions, LLM-guided frameworks bridge the gap between black-box deep learning and human-interpretable spatial reasoning—a critical step toward trustworthy AI in geospatial applications.

\subsection{Cross-City Robustness Analysis}
\label{sec:cross_city}

VPR models employed in our experiments are pre-trained on geographically constrained datasets: CosPlace and EigenPlaces are trained exclusively on SF-XL, while SALAD is trained on GSV-Cities (predominantly containing urban imagery from the U.S. and Europe). This training-inference geographic disparity provides an opportunity to examine cross-city robustness across both common and challenging flooding scenarios.

Under common conditions (Table~S2), all models exhibit substantial performance degradation when tested on Hong Kong compared to San Francisco. VGG16- and ResNet50-based models (CosPlace, EigenPlaces) show approximately 15\% reduction in Recall@1 and 5--10\% reduction in Recall@10. DINOv2-based SALAD demonstrates relatively better cross-city stability, with approximately 10\% reduction in Recall@1 and only 3\% reduction in Recall@10. Under challenging conditions, CosPlace and EigenPlaces maintain similar degradation patterns, while SALAD exhibits markedly smaller performance gaps between cities—particularly the compact 544D variant, which achieves nearly equivalent recall rates on \texttt{sf\_mapi[F]} and \texttt{hk\_mapi[F]} ($\Delta$Recall@1, @10 $< 2\%$).

These results suggest two architectural factors influencing cross-city robustness: (1)~cluster-based aggregation mechanisms (SALAD) appear more resilient to geographic domain shift than GeM pooling methods, potentially due to their ability to capture diverse visual patterns through learned cluster prototypes; and (2)~training on geographically diverse datasets (GSV-Cities) provides moderate improvements over single-city training (SF-XL), though such coverage remains computationally prohibitive at global scale and cannot fully address the long-tail distribution of urban morphologies worldwide.

Notably, VPR-AttLLM maintains consistent improvement margins across both cities despite models' geographic training biases, indicating that LLM-derived semantic priors offer complementary robustness benefits. This raises a compelling research direction: can LLM guidance be extended from post-hoc attention reweighting to deeper integration within feature extraction stages? Current VPR-AttLLM operates at the aggregation layer, modulating existing local features but unable to recover information absent from the original visual descriptors—an intrinsic limitation of post-processing approaches. Future work integrating semantic reasoning earlier in the training pipeline could potentially achieve more fundamental geographic robustness, moving toward VPR systems with global applicability independent of training city distribution. \added{However, achieving global applicability also requires addressing the practical constraints of real-world deployment, particularly computational efficiency during time-critical events.}

\subsection{\added{Operational Utility and Deployment Feasibility}}
\label{sec:computational_efficiency}

\added{To contextualize the practical impact of the absolute recall gains discussed in Section~\ref{fig:experiment_challenging}, it is essential to look beyond the raw retrieval metrics and examine the geospatial distribution of the retrieved data. To illustrate this operational utility, we present a case study using the \texttt{sf\_flood} dataset, selected due to the public availability of San Francisco's official 100-year flood hazard boundaries.}

\begin{figure}[pos=htbp]
    \centering
    \includegraphics[width=\textwidth]{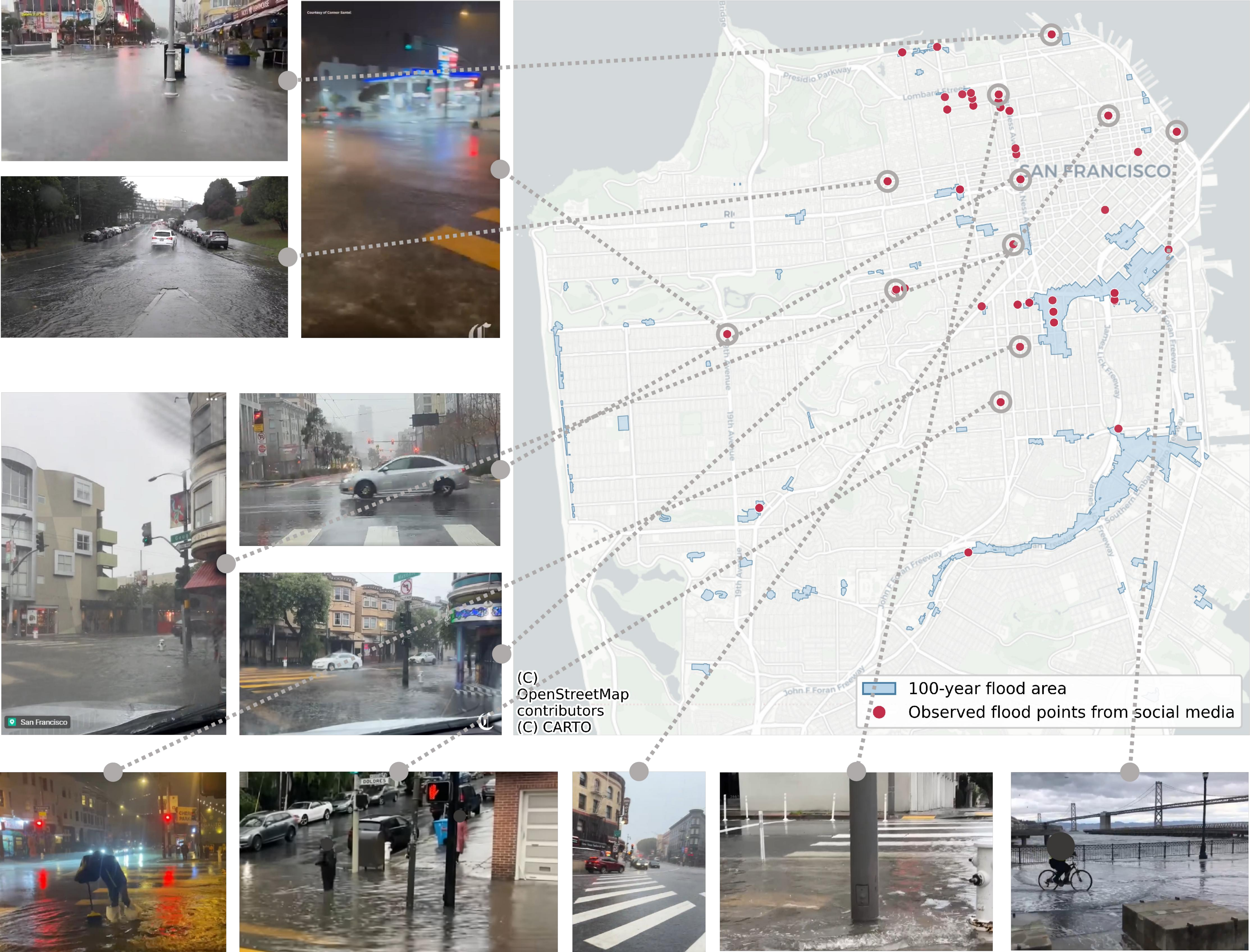}
    \caption{\added{Geospatial distribution of successfully localized crowdsourced imagery in San Francisco, overlaid on the city's official 100-year flood hazard map. The system successfully identifies severe weather and flooding events (red dots) outside anticipated risk zones. Callout images illustrate the localized queries. Note: Points represent geographic clusters; multiple images from nearby or identical intersections from different angles appear as single overlapping points.}}
    \label{fig:sf_floodmap}
\end{figure}

\newcolumntype{S}{>{\raggedright\arraybackslash}p{0.12\textwidth}}

\begin{table}[pos=htbp][H]
    \centering
    \begin{tabularx}{\textwidth}{X S S S S S}
    \toprule
        \textbf{Component} & \textbf{Baseline VPR} & \textbf{Gemini-2.5 (Full)} & \textbf{Gemini-2.5 (Min.)} & \textbf{Qwen3.5-plus (Min.)} & \textbf{Qwen3-vl-8b (Local)} \\
        \midrule
        Total Time / 100 queries
            & 4 s        & 45.7 s    & 37.2 s    & 118.3 s   & 5652.77 s \\
        Avg.\ Time per Query
            & 0.04 s     & 0.46 s    & 0.37 s    & 1.18 s    & 56.53 s \\
        Input Tokens / 100 queries
            & ---        & 106,800   & 44,800    & 79,323    & --- \\
        Total Tokens / 100 queries
            & ---        & 226,053   & 160,520   & 337,176   & --- \\
        Total Cost / 100 queries
            & ---        & \$0.33    & \$0.30    & \$0.13    & --- \\
        Cost per Query
            & ---        & \$0.0033   & \$0.003  & \$0.0013  & --- \\
        \bottomrule
    \end{tabularx}
    \caption{\added{Computational efficiency comparison across VPR pipeline configurations, measured over 100 queries from the \texttt{hk\_flood} dataset. "Full" and "Min." denote full and minimal LLM prompt variants, respectively. Token counts and costs apply only to API-based LLM components. The local Qwen3-VL-8b model was run sequentially on an NVIDIA T4 GPU.}}
    \label{tab:computational_efficiency}
\end{table}

\added{Figure~\ref{fig:sf_floodmap} maps the successfully localized crowdsourced flood images against the official hazard map. As shown, the VPR-AttLLM system successfully geolocates numerous severe weather and flood-related events occurring strictly outside the officially designated high-risk zones. The localized crowdsourced imagery captures a spectrum of conditions—from severe inundation to heavy surface runoff (as illustrated by the callout images). In an operational context, identifying these varying degrees of inundation, provides critical ground-truth data for emergency responders and helps validate dynamic urban hydrological models. While the absolute recall gains (1--8\%) may appear statistically modest, mapping them reveals their high marginal utility. Localizing even a small cluster of these "edge-case" images translates directly to the discovery of unmapped urban infrastructure vulnerabilities.}

\added{To realize this operational value during a time-critical emergency, the system must balance retrieval accuracy with acceptable inference latency. Consequently, we evaluated the deployment feasibility of our framework. Our asymmetric design makes this approach computationally tractable: the LLM processes only the query image, never the massive, offline-indexed reference database.}

\added{Using commercial APIs (e.g., Gemini-2.5-Flash) with parallel request architectures, the per-query latency remains sub-second and the associated economic cost is negligible, as detailed in Table~\ref{tab:computational_efficiency}, less than \$0.5 per 100 queries. While local, open-source deployment (e.g., Qwen3-VL-8B on an NVIDIA T4 GPU) currently requires sequential processing totalling approximately 5652 seconds per 100 queries due to hardware constraints, the API-driven approach demonstrates immediate scalability.}

\added{Crucially, these marginal computational and economic costs must be weighed against the operational reality illustrated in Figure~\ref{fig:sf_floodmap}. Currently, the primary bottleneck in utilizing crowdsourced crisis imagery is the significant labor required for manual expert geo-localization or on-site field surveys, which can take hours per location and incur substantial personnel costs. An additional fraction of a second of computational latency per query is operationally negligible when it directly yields critical ground-truth data—effectively replacing hours of manual geographical investigation by human analysts and accelerating triage and resource dispatch decisions.}

\subsection{Toward Geography-Aware Multimodal VPR Agents}

\change{Beyond immediate operational efficiency,} the broader implication of this work lies in demonstrating that LLMs' geospatial reasoning—acquired from SVI alone—can meaningfully inform visual recognition tasks in real-world deployment scenarios. This capability opens a path toward \textit{geography-aware vision models} that leverage LLMs' implicit knowledge of urban structure (e.g., architectural typologies, street network patterns, landmark hierarchies) to improve cross-domain generalization. For instance, an LLM could dynamically adjust attention priorities based on inferred city context, emphasizing building façades in European historic districts while prioritizing street signage and vertical density cues in Asian megacities. While the current late-fusion strategy treats LLM attention as a post-hoc correction, deeper integration of geospatial reasoning into the learning process holds substantial promise. Future research could explore early-stage integration mechanisms, such as using LLM to guide hard negative mining during contrastive learning, or distilling LLM geospatial priors into lightweight attention modules that can be fine-tuned on target datasets. Such approaches could yield more robust, geography-aware descriptors while maintaining computational feasibility for city-scale deployment.

Beyond visual features, multimodal information present in street-level imagery—particularly text elements such as street signs, storefront names, and directional signage—offers complementary localization cues that remain underutilized in current VPR systems. Although such text regions receive high attention weights in LLM-generated attention maps (Figure~\ref{fig:experiment_analysis1}), they cannot be effectively exploited without corresponding geospatial-aware reference databases. Early works in OCR-augmented VPR~\citep{hazelhoffExploitingStreetlevelPanoramic2014} and recent adavances in LLM-based multimodal reasoning for image geolocalization~\citep{yinLLMenhancedDisasterGeolocalization2025} suggest that fusing visual global descriptors with text embeddings may yield substantial improvements, particularly in urban environments where signage is prevalent and culturally informative. 

Looking forward, LLMs hold potential to serve as \textit{orchestrating agents} that coordinate multiple retrieval modalities—integrating dense street-level imagery with sparse but semantically rich point-of-interest (POI) data, leveraging tool-augmented reasoning for geospatial database queries, and synthesizing evidence across visual, textual, and structural cues. Such agent-based architectures could address challenging real-world localization tasks in emergency response scenarios (e.g., disaster damage assessment from social media imagery) or autonomous navigation in rapidly changing urban environments, where robust place recognition under extreme conditions is not merely an academic exercise but a critical operational requirement. By bridging human-like spatial reasoning with scalable visual retrieval, this research direction promises to transform VPR from a pattern-matching problem into a truly geography-aware, context-sensitive spatial intelligence capability.

\subsection{\added{Ethical Considerations and Broader Impacts}}

\added{The development of advanced Visual Place Recognition (VPR) systems inherently carries dual-use implications that warrant explicit reflection. A primary ethical tension in this work involves the geolocation of crowdsourced imagery, particularly when users may deliberately omit location metadata for privacy reasons. We address this tension across three dimensions.}

\added{First, our system explicitly targets publicly posted social media content during declared crisis events. This imagery is voluntarily shared in a public information space where the poster's implicit intent is to communicate operational or emergency realities. This context is meaningfully distinct from the covert surveillance or tracking of private individuals.}

\added{Second, the fundamental capability to geo-localize street-view imagery from visual content alone is not introduced by this work; it is an established function of VPR systems that have been publicly available and extensively studied for over a decade. Our contribution represents a domain-specific improvement over state-of-art VPR models designed to rescue severely degraded crisis imagery, rather than a qualitative expansion of existing geo-localization capabilities.}

\added{Third, we explicitly acknowledge that any geo-localization technology carries misuse potential, including location inference for stalking or targeted harassment by bad actors. To mitigate these risks, our research paradigm emphasizes operational safeguards. We advocate that the deployment of VPR-AttLLM or similar systems be governed by strict institutional oversight, restricted solely to emergency management contexts with defined data governance policies, and subject to standard responsible disclosure practices. Furthermore, the integration of open-source, locally deployable LLMs (as validated in Section~\ref{sec:result_llm}) allows emergency response agencies to run these models on secure, air-gapped infrastructure, ensuring that sensitive civilian data is not transmitted to proprietary third-party APIs. We encourage the research community to continue developing explicit use-case restrictions and audit frameworks for geo-localization systems applied to crowdsourced content.}

\section{Conclusion} 
\label{sec:conclusion}

This work introduces VPR-AttLLM, a training-free framework that leverages the LLM's geospatial reasoning to enhance visual place recognition under challenging real-world conditions. \change{Evaluated across the morphologically distinct urban environments of San Francisco and Hong Kong, the framework generates semantically informed attention maps that emphasize architecturally distinctive features while suppressing transient visual noise.} This approach achieves consistent recall improvements (1–8\%) across multiple VPR architectures, with the most substantial gains observed in severely flood-degraded imagery. Crucially, as demonstrated through our spatial distribution analysis, these percentage gains translate directly into a critical reduction in absolute physical distance errors, successfully pulling a higher volume of localized images into the operationally useful 0–100 meter range for emergency responders.

\change{The framework operates as a plug-and-play query-side enhancement that requires no model retraining or reference database reprocessing. Furthermore, our rigorous evaluations confirm that VPR-AttLLM is highly robust to prompt phrasing variations and maintains its efficacy when deployed via smaller, open-source Vision-Language Models (e.g., Qwen3-VL-8B). This ensures full reproducibility and enables agencies to deploy the system locally on secure infrastructure, thereby mitigating privacy concerns associated with proprietary API data transmission.}

Beyond immediate technical contributions, this research establishes a methodological pathway for operationalizing urban perception theory within computer vision systems. \change{The framework's cross-domain effectiveness validates that LLM-derived semantic priors encode transferable spatial intelligence that distinctly complements learned visual representations.} As street-level imagery continues to proliferate through crowdsourced platforms, integrating semantic reasoning with visual retrieval represents a foundational step toward next-generation VPR agents capable of human-aligned, context-sensitive localization. By demonstrating that such geography-aware capabilities can emerge from pre-trained language models without task-specific fine-tuning, \change{this work opens scalable, interpretable, and ethically deployable avenues for the rapid geolocalization of crisis imagery in complex built environments.}

\section* {Declaration of generative AI and AI-assisted technologies in the writing process}

During the preparation of this work the author(s) used Claude 4.5 (Sonnet) in order to improve language clarity and readability through grammar refinement and word choice suggestions. After using this tool/service, the author(s) reviewed and edited the content as needed and take(s) full responsibility for the content of the published article.

\section* {Acknowledgements}
The authors gratefully acknowledge partial financial support from the Seed Fund for PI Research – Basic Research (Grant No. 2402101354), University Research Committee, The University of Hong Kong.

\section*{Data and Code Availability}
\change{To facilitate reproducibility and algorithmic transparency, the core implementation code is publicly available via GitHub (detailed in Supplementary Section S5). The raw crowdsourced query images cannot be publicly distributed due to platform privacy and copyright restrictions. However, to fully support verifiable evaluation, all intermediate data structures, the end-to-end evaluation pipeline, and supporting data are available from the corresponding author upon reasonable request.}

\bibliographystyle{apa-doi}

\bibliography{reference}

\clearpage

\setcounter{figure}{0}
\renewcommand{\thefigure}{S\arabic{figure}}
\setcounter{table}{0}
\renewcommand{\thetable}{S\arabic{table}}
\setcounter{section}{0}   
\renewcommand{\thesection}{S\arabic{section}}

\section{Experiment results}
\subsection{Full experiment results}

\begin{table}[pos=htbp]  
    \centering
    \small
    \begin{tabularx}{\textwidth}{>{\raggedright\arraybackslash}p{0.1\textwidth}
                               >{\centering\arraybackslash}p{0.06\textwidth}
                               *{15}{>{\centering\arraybackslash}X}}
		\toprule
		\multirow{2}{=}{\textbf{Model}} & \multirow{2}{=}{$\alpha_{\textbf{LLM}}$} & \multicolumn{3}{c}{\textbf{sf\_flood}} & \multicolumn{3}{c}{\textbf{sf\_v1[F]}} & \multicolumn{3}{c}{\textbf{sf\_mapi[F]}} & \multicolumn{3}{c}{\textbf{hk\_flood}} & \multicolumn{3}{c}{\textbf{hk\_mapi[F]}} \\
		\cmidrule(lr){3-5} \cmidrule(lr){6-8} \cmidrule(lr){9-11} \cmidrule(lr){12-14} \cmidrule(lr){15-17}
		 &  & \textbf{@1} & \textbf{@5} & \textbf{@10} & \textbf{@1} & \textbf{@5} & \textbf{@10} & \textbf{@1} & \textbf{@5} & \textbf{@10} & \textbf{@1} & \textbf{@5} & \textbf{@10} & \textbf{@1} & \textbf{@5} & \textbf{@10} \\
		\midrule
        \multirow{3}{=}{VGG16 (512D, CosPlace)} 
        		 & 0.0 & 38.0 & 44.0 & 47.0 & 43.4 & 53.8 & 58.3 & 47.7 & 59.0 & 63.3 & 24.0 & 37.0 & 43.0 & 36.7 & 49.3 & 53.2 \\
        		\cmidrule(r){2-17}
        		& 0.5 & \textbf{41.0} & \textbf{45.0} & \textbf{50.0} & 44.7 & 55.0 & 59.1 & \textbf{49.5} & \textbf{61.0} & \textbf{64.0} & 23.0 & 39.0 & 47.0 & \textbf{37.5} & \textbf{50.1} & \textbf{54.3} \\
        		\cmidrule(r){2-17}
        		& 1.0 & \textbf{41.0} & \textbf{45.0} & 49.0 & \textbf{44.8} & \textbf{55.2} & \textbf{59.4} & 48.3 & 59.2 & 63.3 & \textbf{25.0} & \textbf{41.0} & \textbf{50.0} & 37.1 & 49.7 & 53.8 \\
        		\midrule
        \multirow{5}{=}{VGG16 (512D, EigenPlaces)} 
        		 & 0.0 & 38.0 & 40.0 & 41.0 & 46.2 & 53.9 & 57.4 & 46.4 & 57.3 & 61.0 & 18.0 & 33.0 & 38.0 & 35.1 & 49.0 & 52.9 \\
        		\cmidrule(r){2-17}
        		& 0.5 & 37.0 & \textbf{42.0} & 42.0 & \textbf{47.5} & 54.4 & 57.6 & 47.4 & \textbf{58.6} & \textbf{62.6} & 18.0 & 35.0 & \textbf{41.0} & 35.8 & 49.9 & \textbf{53.7} \\
        		\cmidrule(r){2-17}
        		& 1.0 & \textbf{39.0} & \textbf{42.0} & \textbf{44.0} & 46.7 & \textbf{54.7} & \textbf{58.4} & \textbf{49.1} & 58.2 & 62.4 & \textbf{19.0} & \textbf{36.0} & \textbf{41.0} & \textbf{36.4} & \textbf{50.0} & 53.6 \\
        		\cmidrule(r){2-17}
        		& QE(k=3) & 38.0 & 40.0 & 41.0 & 45.8 & 51.9 & 54.0 & 46.1 & 54.8 & 57.6 & 18.0 & 31.0 & \textbf{41.0} & 34.9 & 46.3 & 49.9 \\
        		\cmidrule(r){2-17}
        		& QE(k=5) & 38.0 & 40.0 & 41.0 & 44.5 & 53.5 & 54.9 & 46.0 & 55.7 & 59.0 & 18.0 & 32.0 & 37.0 & 33.7 & 48.0 & 51.7 \\
        		\midrule
        \multirow{5}{=}{ResNet50 (512D, CosPlace)} 
        		 & 0.0 & 52.0 & 57.0 & 60.0 & 60.5 & 68.1 & 70.9 & 63.4 & 74.2 & 77.3 & 30.0 & \textbf{45.0} & 47.0 & 44.6 & 58.1 & 63.4 \\
        		\cmidrule(r){2-17}
        		& 0.5 & \textbf{53.0} & 58.0 & 62.0 & \textbf{61.3} & \textbf{68.5} & \textbf{71.1} & 64.8 & 74.8 & 77.8 & 30.0 & \textbf{45.0} & \textbf{48.0} & \textbf{45.9} & \textbf{59.7} & \textbf{64.3} \\
        		\cmidrule(r){2-17}
        		& 1.0 & \textbf{53.0} & \textbf{60.0} & \textbf{64.0} & 61.1 & \textbf{68.5} & 70.7 & \textbf{65.5} & \textbf{75.2} & \textbf{78.3} & 31.0 & 42.0 & \textbf{48.0} & 45.6 & 59.3 & 63.9 \\
        		\cmidrule(r){2-17}
        		& QE(k=3) & 50.0 & 55.0 & 59.0 & 60.3 & 67.1 & 68.5 & 63.7 & 72.3 & 74.9 & \textbf{33.0} & 42.0 & 46.0 & 44.3 & 57.1 & 60.6 \\
        		\cmidrule(r){2-17}
        		& QE(k=5) & 49.0 & 56.0 & 57.0 & 59.8 & 67.3 & 65.8 & 63.4 & 73.1 & 75.4 & 31.0 & 44.0 & 46.0 & 44.8 & 57.1 & 61.6 \\
        		\midrule
        \multirow{3}{=}{ResNet (512D, EigenPlaces)} 
        		 & 0.0 & 51.0 & 60.0 & 65.0 & 65.0 & \textbf{71.6} & 73.4 & 66.4 & 75.3 & 78.5 & \textbf{32.0} & \textbf{43.0} & 47.0 & 47.8 & 60.3 & 64.7 \\
        		\cmidrule(r){2-17}
        		& 0.5 & 52.0 & 60.0 & \textbf{67.0} & 65.5 & \textbf{71.6} & 73.5 & \textbf{67.7} & 75.7 & \textbf{79.2} & 31.0 & \textbf{43.0} & 45.0 & \textbf{48.6} & \textbf{61.4} & \textbf{66.4} \\
        		\cmidrule(r){2-17}
        		& 1.0 & \textbf{53.0} & \textbf{66.0} & \textbf{67.0} & \textbf{65.7} & 71.5 & \textbf{73.7} & 67.3 & \textbf{75.8} & 78.5 & 31.0 & 40.0 & \textbf{48.0} & 47.3 & 60.4 & 65.6 \\
        		\midrule
        \multirow{3}{=}{DINOv2 (8192+256D, SALAD)} 
        		 & 0.0 & 63.0 & 74.0 & 77.0 & 64.1 & 71.1 & 74.3 & 71.1 & 81.3 & 84.3 & \textbf{60.0} & 70.0 & 74.0 & 66.6 & 79.7 & 82.9 \\
        		\cmidrule(r){2-17}
        		& 0.5 & \textbf{64.0} & 74.0 & 77.0 & 64.1 & 71.2 & \textbf{74.7} & \textbf{71.6} & \textbf{81.6} & 84.6 & 59.0 & 71.0 & \textbf{75.0} & \textbf{66.7} & \textbf{79.9} & 83.2 \\
        		\cmidrule(r){2-17}
        		& 1.0 & \textbf{64.0} & \textbf{75.0} & \textbf{79.0} & \textbf{64.3} & \textbf{71.3} & 74.6 & \textbf{71.6} & \textbf{81.6} & \textbf{84.8} & 59.0 & \textbf{72.0} & 74.0 & 66.3 & \textbf{79.9} & \textbf{83.3} \\
        		\midrule
        \multirow{5}{=}{DINOv2 (512+32D, SALAD)} 
        		 & 0.0 & \textbf{50.0} & 59.0 & \textbf{64.0} & 51.4 & 59.5 & 62.9 & 54.6 & 68.4 & \textbf{73.2} & 42.0 & \textbf{58.0} & 60.0 & 53.2 & 70.3 & 75.0 \\
        		\cmidrule(r){2-17}
        		& 0.5 & 48.0 & \textbf{60.0} & 63.0 & \textbf{51.7} & 59.5 & \textbf{63.4} & 54.7 & 68.2 & 72.9 & 42.0 & 56.0 & \textbf{61.0} & 54.6 & 70.5 & 74.8 \\
        		\cmidrule(r){2-17}
        		& 1.0 & 48.0 & \textbf{60.0} & 62.0 & \textbf{51.7} & \textbf{59.8} & \textbf{63.4} & \textbf{54.9} & \textbf{68.7} & 72.8 & \textbf{44.0} & 52.0 & 59.0 & \textbf{54.8} & \textbf{70.9} & \textbf{75.5} \\
        		\cmidrule(r){2-17}
        		& QE(k=3) & 47.0 & 59.0 & 59.0 & 50.6 & 58.0 & 60.4 & 53.7 & 65.9 & 70.2 & 43.0 & 54.0 & 60.0 & 53.4 & 69.5 & 73.5 \\
        		\cmidrule(r){2-17}
        		& QE(k=5) & 48.0 & 59.0 & 60.0 & 50.5 & 59.0 & 61.1 & 54.2 & 67.6 & 70.8 & 42.0 & \textbf{58.0} & \textbf{61.0} & 53.5 & 70.2 & 73.7 \\
        		\bottomrule
    \end{tabularx}
    \captionof{table}{Recall@N performance on across flooding query sets for six VPR architectures under varying LLM attention weights $\alpha_{\text{LLM}} \in \{0.0, 0.5, 1.0\}$, where $\alpha_{\text{LLM}} = 0.0$ represents the unenhanced baseline. Models include CosPlace and EigenPlaces with VGG16 (512D) and ResNet50 (512D) backbones, and SALAD with DINOv2 backbone in full (8448D) and compact (544D) configurations. Query Expansion (QE) baselines with $k \in \{3, 5\}$ are evaluated at $\alpha_{\text{LLM}} = 0.0$ for comparison. Recall@1, @5, and @10 metrics are reported across five challenging scenarios. Best performance for each model-query combination is highlighted in \textbf{bold}.}
	\label{tab:model_performance_challenging}
\end{table} 

\clearpage
\subsection{Experiment results for non-flooding scenarios}

\vspace{10pt}
\noindent
\begin{minipage}{\textwidth}
    \small
    \begin{tabularx}{\textwidth}{>{\raggedright\arraybackslash}p{0.13\textwidth}
                                   >{\centering\arraybackslash}p{0.05\textwidth}
                                   *{9}{>{\centering\arraybackslash}p{0.06\textwidth}}}
		\toprule
		\multirow{2}{=}{\textbf{Model}} & \multirow{2}{=}{$\alpha_{\textbf{LLM}}$} & \multicolumn{3}{c}{\textbf{sf\_v1}} & \multicolumn{3}{c}{\textbf{sf\_mapi}} & \multicolumn{3}{c}{\textbf{hk\_mapi}} \\
		\cmidrule(lr){3-5} \cmidrule(lr){6-8} \cmidrule(lr){9-11}
		 &  & \textbf{@1} & \textbf{@5} & \textbf{@10} & \textbf{@1} & \textbf{@5} & \textbf{@10} & \textbf{@1} & \textbf{@5} & \textbf{@10} \\
		\midrule
        \multirow{5}{=}{VGG16 (512D, EigenPlaces)} 
        		 & 0.0 & \textbf{70.4} & \textbf{78.7} & 81.6 & 72.7 & 82.4 & 85.2 & \textbf{58.8} & 72.9 & 77.6 \\
        		\cmidrule(r){2-11}
        		& 0.5 & 70.2 & \textbf{78.7} & \textbf{81.7} & \textbf{73.6} & \textbf{82.8} & \textbf{85.5} & 58.3 & \textbf{73.7} & \textbf{78.2} \\
        		\cmidrule(r){2-11}
        		& 1.0 & 70.2 & 78.2 & 81.1 & 73.5 & 82.4 & 85.0 & 58.0 & 73.5 & 78.0 \\
        		\cmidrule(r){2-11}
        		& QE(k=3) & 69.9 & 76.3 & 78.8 & 72.6 & 80.5 & 82.2 & 58.0 & 72.0 & 75.5 \\
        		\cmidrule(r){2-11}
        		& QE(k=5) & 70.0 & 78.6 & 79.6 & 73.5 & 81.6 & 82.8 & 58.1 & 72.2 & 75.3 \\
        		\midrule
        \multirow{5}{=}{ResNet50 (512D, CosPlace)} 
        		 & 0.0 & \textbf{78.3} & \textbf{84.6} & 86.8 & 82.2 & 89.8 & 91.3 & \textbf{67.2} & \textbf{81.7} & \textbf{85.1} \\
        		\cmidrule(r){2-11}
        		& 0.5 & 78.0 & 84.5 & \textbf{87.0} & \textbf{82.5} & 89.9 & \textbf{91.6} & 67.1 & \textbf{81.7} & \textbf{85.1} \\
        		\cmidrule(r){2-11}
        		& 1.0 & 78.1 & 84.1 & 86.9 & 82.3 & \textbf{90.1} & \textbf{91.6} & 66.6 & 81.5 & 84.6 \\
        		\cmidrule(r){2-11}
        		& QE(k=3) & 77.4 & 83.7 & 85.3 & 81.9 & 88.5 & 90.0 & 67.1 & 80.8 & 83.6 \\
        		\cmidrule(r){2-11}
        		& QE(k=5) & 77.4 & 84.2 & 85.1 & 82.4 & 89.2 & 90.6 & 67.0 & 80.9 & 83.8 \\
        		\midrule
        \multirow{5}{=}{DINOv2 (8448D, SALAD)} 
        		 & 0.0 & 88.7 & \textbf{93.5} & 94.4 & \textbf{85.6} & 93.2 & 94.8 & 76.1 & \textbf{89.4} & 91.4 \\
        		\cmidrule(r){2-11}
        		& 0.5 & \textbf{88.8} & 93.4 & 94.4 & 85.0 & \textbf{93.5} & 94.8 & 76.1 & \textbf{89.4} & \textbf{91.5} \\
        		\cmidrule(r){2-11}
        		& 1.0 & \textbf{88.8} & 93.4 & \textbf{94.5} & 85.0 & \textbf{93.5} & \textbf{95.1} & \textbf{76.2} & 89.3 & \textbf{91.5} \\
        		\cmidrule(r){2-11}
        		& QE(k=3) & 80.7 & 88.0 & 88.7 & 76.0 & 86.1 & 88.4 & 71.1 & 85.7 & 87.6 \\
        		\cmidrule(r){2-11}
        		& QE(k=5) & 80.8 & 88.9 & 89.6 & 75.0 & 86.7 & 88.7 & 71.3 & 86.1 & 87.9 \\
        		\bottomrule
    \end{tabularx}
    \captionof{table}{Recall@N performance on common query sets across six VPR architectures with varying LLM attention weights $\alpha_{\text{LLM}} \in \{0.0, 0.5, 1.0\}$, where $\alpha_{\text{LLM}} = 0.0$ represents the baseline. Query Expansion (QE) results with $k \in \{3, 5\}$ top retrievals are included for comparison. Best performance for each model-query combination is shown in \textbf{bold}. VPR-AttLLM maintains stable performance with modest improvements on heterogeneous Mapillary queries (\texttt{sf\_mapi}, \texttt{hk\_mapi}), while QE consistently degrades baseline performance across all scenarios.}
	\label{tab:model_performance_common}
\end{minipage}
\vspace{15pt}

\clearpage
\subsection{Experiment results for reversed city used in prompt}
\label{sec:supple_reverse}

\vspace{10pt}
\noindent
\begin{minipage}{\textwidth}
    \small
    \begin{tabularx}{\textwidth}{>{\raggedright\arraybackslash}p{0.1\textwidth}
                               >{\centering\arraybackslash}p{0.06\textwidth}
                               *{15}{>{\centering\arraybackslash}X}}
		\toprule
		\multirow{2}{=}{\textbf{Model}} & \multirow{2}{=}{$\alpha_{\textbf{LLM}}$} & \multicolumn{3}{c}{\textbf{sf\_flood}} & \multicolumn{3}{c}{\textbf{sf\_v1[F]}} & \multicolumn{3}{c}{\textbf{sf\_mapi[F]}} & \multicolumn{3}{c}{\textbf{hk\_flood}} & \multicolumn{3}{c}{\textbf{hk\_mapi[F]}} \\
		\cmidrule(lr){3-5} \cmidrule(lr){6-8} \cmidrule(lr){9-11} \cmidrule(lr){12-14} \cmidrule(lr){15-17}
		 &  & \textbf{@1} & \textbf{@5} & \textbf{@10} & \textbf{@1} & \textbf{@5} & \textbf{@10} & \textbf{@1} & \textbf{@5} & \textbf{@10} & \textbf{@1} & \textbf{@5} & \textbf{@10} & \textbf{@1} & \textbf{@5} & \textbf{@10} \\
		\midrule
        \multirow{5}{=}{VGG16 (512D, EigenPlaces)} 
        		 & 0.0 & 38.0 & 40.0 & 41.0 & 46.2 & 53.9 & 57.4 & 46.4 & 57.3 & 61.0 & 18.0 & 33.0 & 38.0 & 35.1 & 49.0 & 52.9 \\
        		\cmidrule(r){2-17}
        		& 0.5 & 37.0 & \textbf{42.0} & 42.0 & \textbf{47.5} & 54.4 & 57.6 & 47.4 & 58.6 & \textbf{62.6} & 18.0 & 35.0 & 41.0 & 35.8 & 49.9 & 53.7 \\
        		\cmidrule(r){2-17}
        		& 1.0 & \textbf{39.0} & \textbf{42.0} & \textbf{44.0} & 46.7 & \textbf{54.7} & \textbf{58.4} & \textbf{49.1} & 58.2 & 62.4 & 19.0 & 36.0 & 41.0 & \textbf{36.4} & \textbf{50.0} & 53.6 \\
        		\cmidrule(r){2-17}
        		& 0.5 (R) & \textbf{39.0} & 41.0 & 42.0 & 47.0 & 54.4 & 57.7 & 47.1 & \textbf{59.2} & 62.3 & 18.0 & 32.0 & 41.0 & 35.4 & 49.2 & \textbf{53.8} \\
        		\cmidrule(r){2-17}
        		& 1.0 (R) & 36.0 & 41.0 & 42.0 & 46.5 & 54.2 & 57.7 & 48.6 & 58.2 & 62.2 & \textbf{20.0} & \textbf{37.0} & \textbf{42.0} & 34.5 & 47.6 & 51.7 \\
        		\midrule
        \multirow{5}{=}{ResNet50 (512D, CosPlace)} 
        		 & 0.0 & 52.0 & 57.0 & 60.0 & 60.5 & 68.1 & 70.9 & 63.4 & 74.2 & 77.3 & 30.0 & 45.0 & 47.0 & 44.6 & 58.1 & 63.4 \\
        		\cmidrule(r){2-17}
        		& 0.5 & \textbf{53.0} & 58.0 & 62.0 & \textbf{61.3} & \textbf{68.5} & \textbf{71.1} & 64.8 & 74.8 & 77.8 & 30.0 & 45.0 & 48.0 & \textbf{45.9} & 59.7 & 64.3 \\
        		\cmidrule(r){2-17}
        		& 1.0 & \textbf{53.0} & \textbf{60.0} & \textbf{64.0} & 61.1 & \textbf{68.5} & 70.7 & \textbf{65.5} & \textbf{75.2} & \textbf{78.3} & 31.0 & 42.0 & 48.0 & 45.6 & 59.3 & 63.9 \\
        		\cmidrule(r){2-17}
        		& 0.5 (R) & 52.0 & 59.0 & 62.0 & 61.0 & \textbf{68.5} & 70.9 & 65.4 & 74.5 & 78.0 & 34.0 & \textbf{46.0} & 48.0 & 45.3 & \textbf{59.9} & \textbf{64.5} \\
        		\cmidrule(r){2-17}
        		& 1.0 (R) & \textbf{53.0} & 59.0 & \textbf{64.0} & 61.1 & 68.2 & 70.8 & \textbf{65.5} & 74.2 & 77.2 & \textbf{35.0} & 44.0 & \textbf{51.0} & 45.7 & 58.3 & 63.1 \\
        		\midrule
        \multirow{5}{=}{DINOv2 (544D, SALAD)} 
        		 & 0.0 & \textbf{50.0} & 59.0 & \textbf{64.0} & 51.4 & 59.5 & 62.9 & 54.6 & 68.4 & 73.2 & 42.0 & \textbf{58.0} & 60.0 & 53.2 & 70.3 & 75.0 \\
        		\cmidrule(r){2-17}
        		& 0.5 & 48.0 & \textbf{60.0} & 63.0 & \textbf{51.7} & 59.5 & \textbf{63.4} & 54.7 & 68.2 & 72.9 & 42.0 & 56.0 & 61.0 & 54.6 & 70.5 & 74.8 \\
        		\cmidrule(r){2-17}
        		& 1.0 & 48.0 & \textbf{60.0} & 62.0 & \textbf{51.7} & \textbf{59.8} & \textbf{63.4} & \textbf{54.9} & 68.7 & 72.8 & \textbf{44.0} & 52.0 & 59.0 & \textbf{54.8} & \textbf{70.9} & \textbf{75.5} \\
        		\cmidrule(r){2-17}
        		& 0.5 (R) & 49.0 & \textbf{60.0} & 63.0 & 51.2 & 59.6 & \textbf{63.4} & \textbf{54.9} & 68.8 & 73.0 & 42.0 & 57.0 & 63.0 & 54.2 & 70.5 & 74.7 \\
        		\cmidrule(r){2-17}
        		& 1.0 (R) & 48.0 & 59.0 & \textbf{64.0} & 51.5 & 59.7 & 63.3 & 54.3 & \textbf{69.9} & \textbf{73.3} & 41.0 & 56.0 & \textbf{65.0} & 54.2 & 70.2 & 74.3 \\
        		\bottomrule
    \end{tabularx}
    \captionof{table}{Ablation study of geographic context correctness (correct vs.\ reversed city context) on Recall@N across all query sets and backbone architectures, with LLM attention weights $\alpha_{\text{LLM}} \in \{0.5, 1.0\}$ and baseline at $\alpha_{\text{LLM}} = 0.0$. Reversed context, denoted by (R), consistently incurs a modest roughly $1\%$ Recall@1 degradation relative to correct context, yet remains above the baseline across nearly all model--query configurations, demonstrating robustness to imperfect geographic metadata. Best performance for each model--query combination is shown in \textbf{bold}.}
\end{minipage}

\clearpage
\section{Results analysis}
\begin{figure}[pos=htbp]
    \centering
    \includegraphics[width=0.5\textwidth]{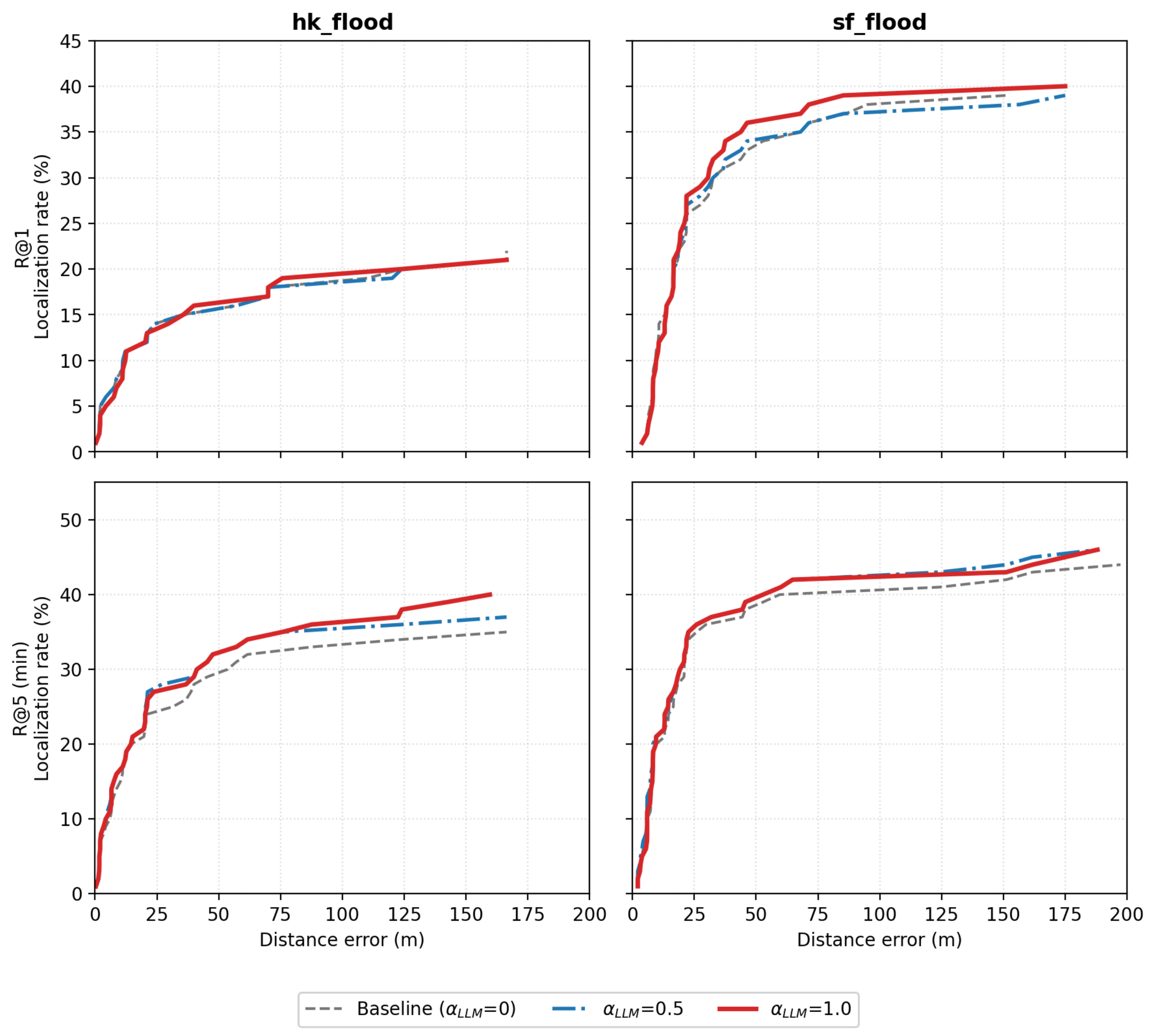}
    \caption{\added{CDF of localization distance errors for datasets hk\_flood and sf\_flood under R@1 (top) and R@5-min (bottom) metrics using EigenPlaces (VGG16). Higher $\alpha_{\text{LLM}}$ values consistently shift curves leftward, indicating improved localization accuracy within tight spatial thresholds relative to the unenhanced baseline.}}
    \label{fig:experiment_cdf_eigenplaces} 
\end{figure}

\section {Prompts used in this study}
\label{sec:supple_prompt}

\subsection{Prompts for generating attention weights from SVI}

\subsubsection{Full structured prompts}
* VPR Geo-localization: Scene Composition Weighting

**TASK:** 
Your goal is to determine if an image requires spatial weighting for Visual Place Recognition (VPR) and, if so, to provide those weights. Follow this two-step decision process strictly.

STEP 1: DECIDE THE SCENE TYPE

First, classify the image into one of two categories. This is the most critical step.

A) 'Single Landmark' Image: The image is dominated by one primary, cohesive structure.
Criteria: This includes focused shots of a single building's facade, a prominent monument, a statue, or a bridge, etc. The object acts as a single, unambiguous identifier.
Crucial Rule: Even if this single landmark has unique details (e.g., intricate windows, a brand name, specific textures), it must be treated as one whole entity. Do not segment a single building into parts. Applying different weights to parts of one building is incorrect and adds noise.
ACTION: If the image is a 'Single Landmark', STOP. Your entire and only output must be the single string: "None"

B) 'Complex Scene' Image: The image contains multiple, distinct, and geographically significant objects that contribute differently to localization.
Criteria: This includes street corners with several buildings, plazas with fountains and surrounding structures, or street views with varied storefronts and unique street furniture. The combination of these separate objects defines the location.
ACTION: If the image is a 'Complex Scene', PROCEED to Step 2 to generate weights.

STEP 2: GENERATE WEIGHTS (FOR 'COMPLEX SCENE' IMAGES ONLY)
If and only if you identified the image as a 'Complex Scene' in Step 1, identify 3-8 key regions and assign weights based on their uniqueness for localization in the city.
The city is {city}, and the spatial uniqueness is the uniqueness of the region in the whole city.

**COORDINATE SYSTEM:**
Image coordinates range from (0.0, 0.0) at top-left to (1.0, 1.0) at bottom-right
Center coordinates represent the focal point of each distinctive region

**WEIGHTING LOGIC:**
**HIGH WEIGHTS (1.6-2.0):** Spatially unique in {city}
- Rare architectural details, unusual building features
- Distinctive facades, unique structural elements
- Sharp, detailed building textures/patterns
- Uncommon urban installations or specialized infrastructure

**MEDIUM WEIGHTS (1.0-1.5):** Moderately distinctive in {city}
- Well-detailed standard buildings with clear architectural features
- Street-level infrastructure with visible specificity
- Clear signage, distinctive storefronts

**LOW WEIGHTS (0.3-0.9):** Common elements in {city}
- Typical building styles (abundant citywide)
- Generic street infrastructure, standard traffic elements

**MINIMAL WEIGHTS (0.0-0.2):** Non-localizable in {city}
- Sky, vegetation, generic pavement
- People, vehicles, temporary objects
- Blurred or indistinct regions

**SELECTION STRATEGY:**
Focus on 3-8 most important regions that deviate significantly from default weight (1.0)
Prioritize regions with highest uniqueness (>1.5) and lowest uniqueness (<0.5)
Skip regions that would receive approximately default weight (0.9-1.1)

\begin{verbatim}
**OUTPUT FORMAT:**
[
    {
        "center": [x_coord, y_coord],
        "weight": weight_value,
        "reasoning": "brief_description"
    }
]
\end{verbatim}

\subsubsection{\added{Minimal prompts}}

Analyze the image for VPR spatial weighting in {city}.

Identify 3-8 key regions and assign weights based on uniqueness for localization in {city}:
- 1.6–2.0: Spatially unique
- 1.0–1.5: Moderately distinctive
- 0.3–0.9: Common
- 0.0–0.2: Non-localizable (sky, vegetation, people, vehicles)

Skip regions near weight 1.0. Use coordinates (0.0, 0.0) top-left to (1.0, 1.0) bottom-right.
\begin{verbatim}
[
  {
      "center": [x_coord, y_coord],
      "weight": weight_value,
      "reasoning": "brief_description"
  }
]
\end{verbatim}

\subsection {Prompts for simulating flooding queries}
\label{sec:supple_sim}
Simulate a realistic slight inundation at this location. Strictly preserve all original buildings and infrastructure. Maintain the original camera viewpoint. Add realistic floodwaters and overcast, stormy weather. The final image should look like a photo taken in the rain, with imperfections like a few water droplets and a smear across a portion of the lens.

\clearpage
\section{Failure case analysis}

\begin{figure}[pos=htbp]
    \centering
    \includegraphics[width=\textwidth]{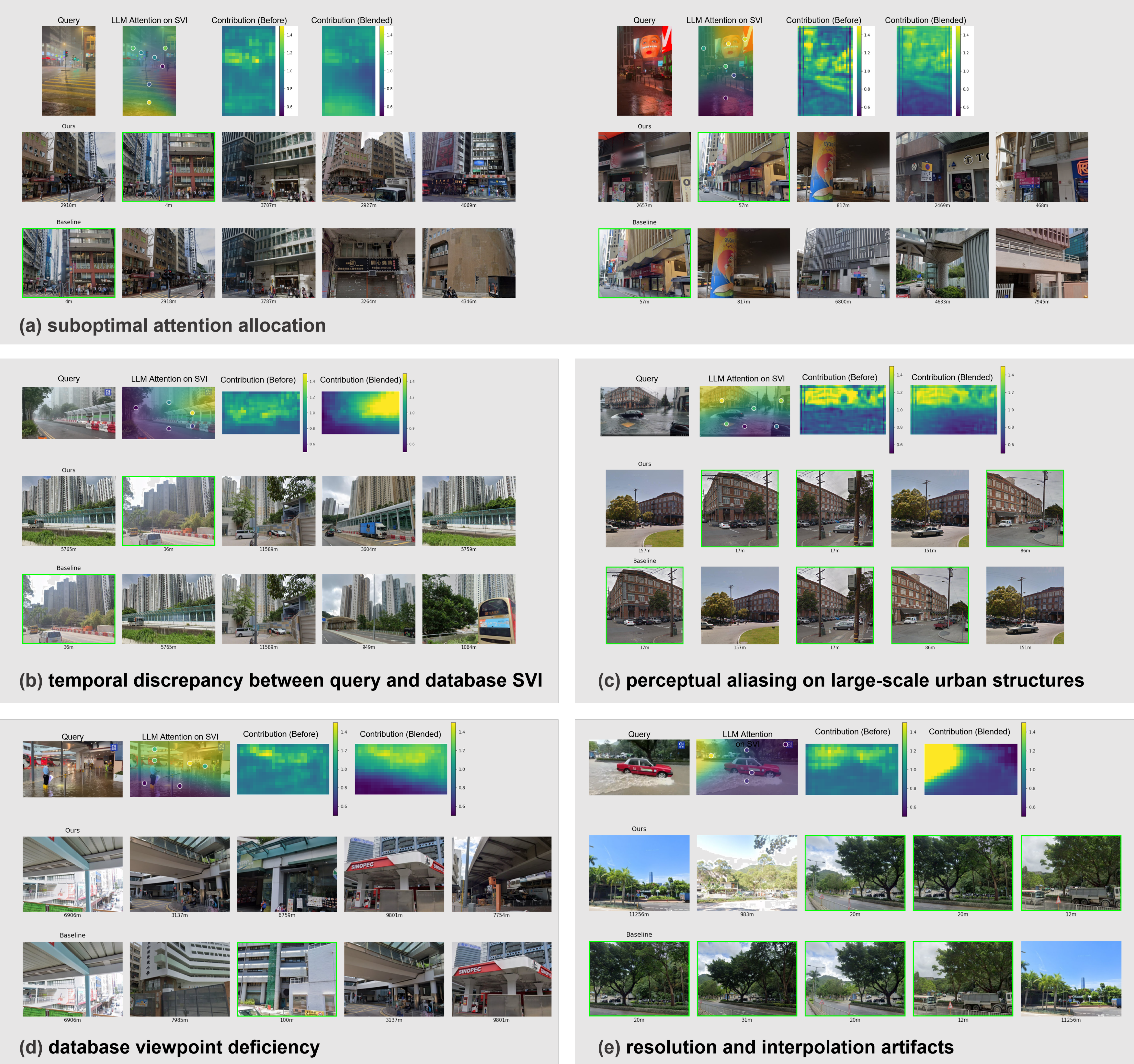}
    \caption{\added{\textbf{Qualitative analysis of representative failure modalities in LLM-guided geo-localization.} Each panel (a–e) follows a consistent layout: the top row displays the query image alongside heatmaps illustrating the LLM attention maps and their blending effects on spatial feature contributions; the middle row shows the top-5 retrieval results from the proposed VPR-AttLLM, while the bottom row presents the top-5 results from the baseline model. Green bounding boxes denote correct ground-truth matches within the accepted distance threshold. These cases illustrate how performance is impacted by the specific environmental and algorithmic constraints labeled in each sub-figure, such as temporal shifts, viewpoint deficiencies, and urban aliasing.}
}
    \label{fig:S1}
\end{figure}

\section{Implementation Code}
The core implementation code for this work is publicly available at the following GitHub repository:

\url{https://github.com/fengyixu/VPR_AttLLM}

\end{document}